\documentclass[10pt,twocolumn,letterpaper]{article}
\pdfoutput=1
\usepackage[pagenumbers]{cvpr} %

\usepackage{graphicx}
\usepackage{booktabs}
\usepackage{amsmath}
\usepackage{amssymb}
\usepackage{multirow}
\usepackage{xcolor}
\usepackage{tabularx}
\usepackage{colortbl}  %
\usepackage{array}
\usepackage{siunitx} %
\usepackage{tcolorbox}  %
\usepackage{listings}
\usepackage{mathrsfs}
\usepackage{wrapfig}
\usepackage{upgreek}
\usepackage{makecell}
\usepackage{vcell}
\usepackage{mwe} %
\usepackage{calc} %
\usepackage{pifont} %

\usepackage{marvosym}   %

\usepackage{epigraph}
\newlength\savewidth\newcommand\shline{\noalign{\global\savewidth\arrayrulewidth
		\global\arrayrulewidth 1pt}\hline\noalign{\global\arrayrulewidth\savewidth}}

\newcommand{\yes}{\color{blue}{\ding{51}}}
\newcommand{\no}{\color{red}{\ding{55}}}
\definecolor{Gray}{gray}{0.85}
\newcommand{\method}{\mbox{{AnyBimanual}}\xspace}

\definecolor{best}{rgb}{0.96, 0.57, 0.58}
\definecolor{second}{rgb}{0.98, 0.78, 0.57}
\definecolor{third}{rgb}{1.0, 1.0, 0.56}

\newcommand{\ourcolor}{gray!9}

\definecolor{cvprblue}{rgb}{0.21,0.49,0.74}
\usepackage[pagebackref,breaklinks,colorlinks,allcolors=cvprblue]{hyperref}

\def\paperID{4528} %
\def\confName{CVPR}
\def\confYear{2025}

\title{AnyBimanual: Transferring Unimanual Policy for General Bimanual Manipulation}

\author{
    Guanxing Lu$^{\ast,1}$, 
    Tengbo Yu$^{\ast,1}$,
    Haoyuan Deng$^{2}$,
    Season Si Chen$^{1}$,
    Yansong Tang$^{\dagger, 1}$,
    Ziwei Wang$^{2}$\\
    $^1$Tsinghua Shenzhen International Graduate School, Tsinghua University\\
    $^2$School of Electrical and Electronic Engineering, Nanyang Technological University \\
    {\tt\small \{lgx23@mails.,ytb23@mails.,season.chen@,tang.yansong@\}sz.tsinghua.edu.cn}\\
    {\tt\small \{E230112@e.,ziwei.wang@\}ntu.edu.sg}\\
    {\small \url{https://anybimanual.github.io/}
    }
}

\begin{document}

\maketitle
\begin{abstract}

Performing general language-conditioned bimanual manipulation tasks is of great importance for many applications ranging from household service to industrial assembly. However, collecting bimanual manipulation data is expensive due to the high-dimensional action space, which poses challenges for conventional methods to handle general bimanual manipulation tasks. In contrast, unimanual policy has recently demonstrated impressive generalizability across a wide range of tasks because of scaled model parameters and training data, which can provide sharable manipulation knowledge for bimanual systems. To this end, we propose a plug-and-play method named \textbf{AnyBimanual}, which transfers pretrained unimanual policy to general bimanual manipulation policy with few bimanual demonstrations. Specifically, we first introduce a skill manager to dynamically schedule the skill representations discovered from pretrained unimanual policy for bimanual manipulation tasks, which linearly combines skill primitives with task-oriented compensation to represent the bimanual manipulation instruction. To mitigate the observation discrepancy between unimanual and bimanual systems, we present a visual aligner to generate soft masks for visual embedding of the workspace, which aims to align visual input of unimanual policy model for each arm with those during pretraining stage. AnyBimanual shows superiority on 12 simulated tasks from RLBench2 with a sizable 17.33\% improvement in success rate over previous methods. Experiments on 9 real-world tasks further verify its practicality with an average success rate of 84.62\%.

\end{abstract}

% Extensive results on 12 simulated and 9 real-world tasks indicate the superiority of AnyBimanual with an absolute improvement of 12.67\% on average success rate compared with previous state-of-the-art methods, .
% Extensive results on 12 simulated tasks indicate the superiority of AnyBimanual, with an absolute improvement of 12.67\% on average success rate compared with previous state-of-the-art methods. 
% with only 20\% percentage of demonstrations compared with previous state-of-the-art methods.

% \epigraph{``Life can only be understood backwards; but it must be lived forwards.''}{--- Søren Kierkegaard}

% \epigraph{`A complex system that works is invariably found to have evolved from a simple system that worked. ''}{--- Gall's Law}

% \epigraph{`A complex system that works is ... evolved from a simple system that worked. ''}{--- Gall's Law}

% A complex system designed from scratch never works and cannot be patched up to make it work. You have to start over with a working simple system.

% Take a deep breath, I want you to act as an experienced academician. I'm writing a scientific paper for CVPR conference. Here is one part of my draft, please revise it to make it more fluent and compatible with CVPR conference. Only use simple words and expressions. This is very important to my career. """ """

\section{Introduction}
\label{sec:intro}
% Automated language-conditioned bimanual manipulation tasks has been highly demanded by a wide spectrum of applications, 
Bimanual systems play an important role in robotic manipulation due to the high capacity of completing diverse tasks in household service \citep{zhang2024empoweringembodiedmanipulationbimanualmobile}, robotic surgery \citep{10149474}, and component assembly in factories \citep{buhl2019dual}. Compared to unimanual systems, bimanual systems enlarge the workspace and are able to handle more complex manipulation tasks by stabilizing the target with one arm and interacting with that using another arm \citep{grannen2023stabilizeactlearningcoordinate, liu2024voxactbvoxelbasedactingstabilizing}. Even for the tasks that unimanual policies can handle, bimanual systems are often more efficient because multiple action steps can be simultaneously accomplished \citep{grotz2024peract2benchmarkinglearningrobotic}. Since modern robotic applications require the robot to interact with different tasks and objects, it is desirable to design a generalizable policy model for bimanual manipulation.

\begin{figure}[t]
    \centering
    \includegraphics[width=0.47\textwidth]{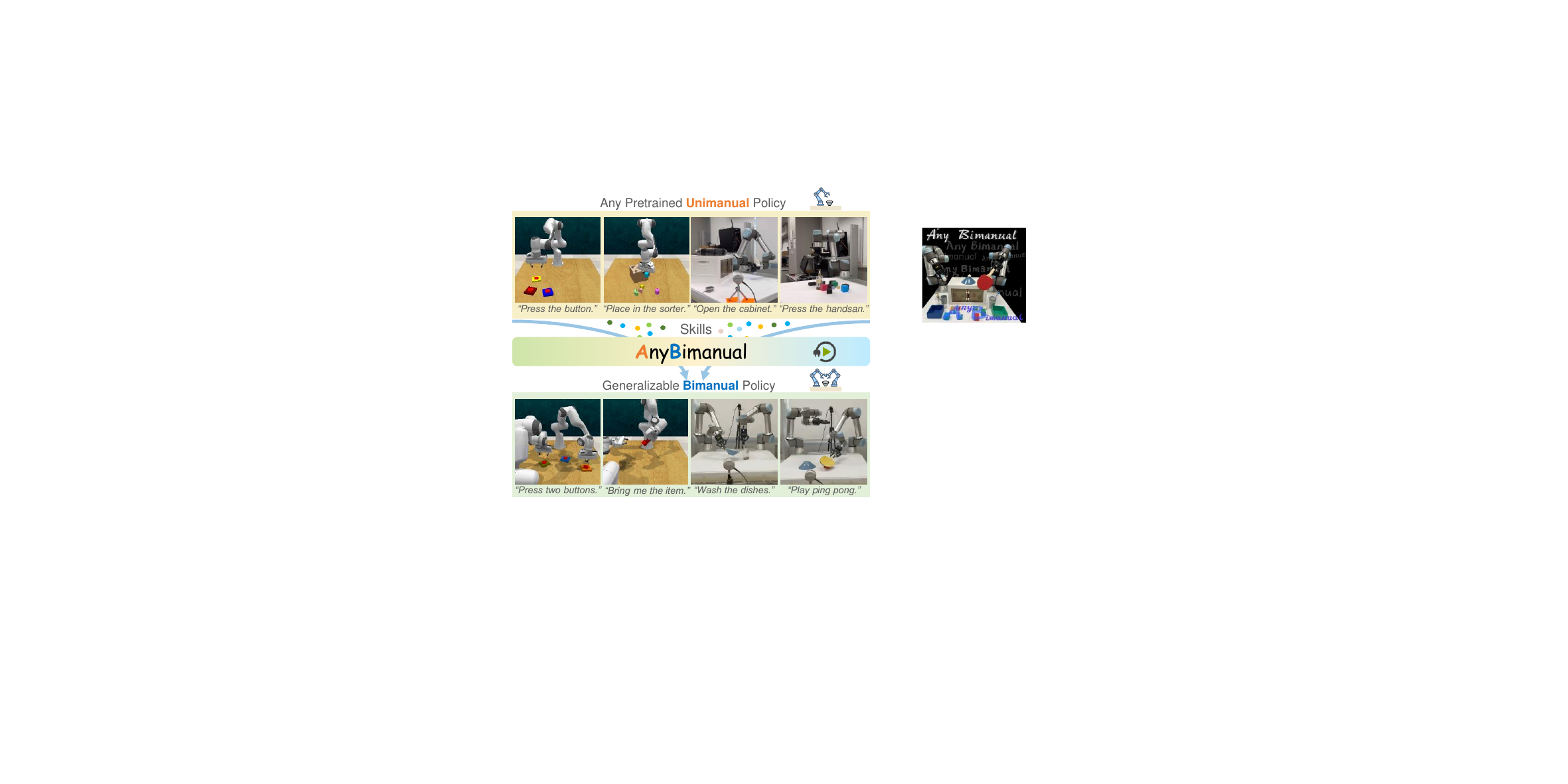}
    \caption{\small \textbf{AnyBimanual} enables plug-and-play transferring from pretrained unimanual policies to bimanual manipulation policy, which preserves the generalizability with the proposed skill scheduling framework. 
    % Right: \method surpasses the state-of-the-art general bimanual manipulation agent \citep{grotz2024peract2benchmarkinglearningrobotic} by a large improvement of 12.67\%.}
    % and XX\% in simualted and real-world tasks, respectively.}
    }
    \label{fig:teaser}
    % \vspace{-0.1cm}
\end{figure}

% single task
%prior works ppresent to define some atomic movements like lift and place, which are utilized to solve unseen tasks via hierarchical planning. Riding into the era of deep learning, 

%% VLM, LLM
To enhance the generalization ability of the manipulation agent, prior works present to leverage the high-level reasoning and semantic understanding capabilities of foundation models like Large Language Models (LLMs) and Vision Language Models (VLMs) to breakdown tasks into executable sub-tasks that can be solved by external low-level controllers \citep{huang2023voxposercomposable3dvalue, gao2024dagplangeneratingdirectedacyclic, liu2023llmpempoweringlargelanguage, joublin2023copalcorrectiveplanningrobot, gbagbe2024bivlavisionlanguageactionmodelbasedbimanual}, which thus struggles with contact-rich tasks that require complex and precise low-level motion.
%% unimanual RFM, but lack data
To generalize to contact-rich tasks, recent methods  \citep{kim2024openvlaopensourcevisionlanguageactionmodel, octomodelteam2024octoopensourcegeneralistrobot, ahn2024autortembodiedfoundationmodels} tend to learn robotic foundation models directly from large-scale teleoperation data  \citep{khazatsky2024droidlargescaleinthewildrobot, embodimentcollaboration2024openxembodimentroboticlearning, walke2024bridgedatav2datasetrobot}, which has shown impressive generalizability across various unimanual tasks.
%To replicate this success to bimanual settings, many methods [ALOHA, mobile ALOHA] are proposed to train the policy model with large-scale expert demonstrations by imtiation learning.
However, bimanual demonstrations are extremely expensive to acquire in real-world, which often need specialized teleoperation systems with additional sensors and fine-grained calibration with high human laborer cost \citep{tony2023aloha, fu2024mobilealohalearningbimanual, ding2024bunnyvisionprorealtimebimanualdexterous, cheng2024opentelevisionteleoperationimmersiveactive, wang2024dexcapscalableportablemocap, wu2024gellogenerallowcostintuitive}. 
%% Skill learning, but predefined is not flexible 
To address this challenge, recent methods aim to simplify the learning budget by exploiting human inductive bias like parameterized atomic movements that detail the position and rotation \citep{chitnis2020efficient, Gong_2023} or assigning stabilizing and acting roles for each arm \citep{liu2024voxactbvoxelbasedactingstabilizing, varley2024embodiedaiarmszeroshot, gao2024bikvilkeypointsbasedvisualimitation}, thereby reducing the need for extensive expert data.
% to handle different tasks by combining different atomic movements. 
% discrimitive
Nevertheless, shareable atomic movements and fixed cooperation patterns struggle to generalize across different bimanual manipulation tasks, which limits the deployment scenario of these classes of methods. 

% cooperation patterns
% fixed movement and direction, collaboration pattern
% are often hard to specified even by human users,

%On the opposite, unimanual demonstrations are relatively easy to obtain by a single VR controller [PerAct, GNFactor], space mouse [xx], joystick [xx], or even keyboard [DP3].

In this paper, we propose a plug-and-play module named \method that transfers any pre-trained unimanual policy to bimanual manipulation policy with limited demonstrations. 
Since unimanual policy model \citep{shridhar2022perceiveractormultitasktransformerrobotic, ke20243ddiffuseractorpolicy} has demonstrated impressive generalization ability across tasks due to the large model sizes and numerous training demonstrations, we realize high generalizability across diverse language-conditioned bimanual manipulation tasks by mining and transferring the commonsense knowledge in pretrained unimanual policies. More specifically, we first introduce a skill manager that dynamically schedules discovered skill representations.
Skill representations are formed by skill primitives that store shareable manipulation knowledge across embodiments, with task-oriented importance weights and compensation.
% from the pretrained unimanual policy. 
% Skill primitives demonstrates the shareable manipulation knowledge across embodiments, which are combined with importance weights and task-oriented compensation. 
To enhance the transferability of the pretrained unimanual policy in bimanual manipulation tasks, the observation discrepancy between unimanual and bimanual systems should be minimized. We propose a voxel aligner to generate spatial soft masks to highlight relevant visual clues for different arms, whose goal is to align the visual input of the unimanual policy model for each arm with those during the pretraining stage. 
% \Cref{fig:toy} shows an example of bimanual manipulation policy composition from two unimanual policy models for the left and right arms.
%decomposes the 3D voxel input for each arm by tuning a spatial soft mask, where a regularization term is included to ensure that each arm focuses on different areas of interest. The predicted voxel input and skill embedding are then passed through two multi-modal unimanual multi-task policies that represent the left and right arm to obtain the optimal next end-effector pose. 
We evaluate \method on a comprehensive task suite composed of $12$ simulated tasks from RLBench2~\citep{grotz2024peract2benchmarkinglearningrobotic} and $9$ real-world tasks, where our method surpasses the previous state-of-the-art method by a large margin. The contributions are summarized as follows:

\begin{itemize}
    \item We propose a model-agnostic plug-and-play framework named \method that transfers an arbitrary pretrained unimanual policy to generalizable bimanual manipulation policy with limited bimanual demonstrations.

    % with linear representation weights and error compensation
    \item  We introduce a skill manager to dynamically schedule skill representations for unimanual policy transferring and a visual aligner to mitigate the observation discrepancy between unimanual and bimanual systems for transferability enhancement.

    \item We conduct extensive experiments of $12$ tasks from RLBench2 and $9$ tasks from the real world. The results demonstrate that our method achieves a higher success rate than the state-of-the-art methods.
    % with less demonstrations.
    
\end{itemize}

% Assumption: All bimanual manipulation tasks can be solved by correctly scheduling unimanual manipulation subtasks [xx].

% Take a deep breath, I want you to act as an experienced academician. I'm writing a scientific paper for CVPR conference. Here is the related work of my draft, please revise it to make it more fluent and compatible with CVPR conference. Make all expression logic. Only use simple words and expressions. This is very important to my career. """ """

\section{Related Work}
\label{sec:related_work}

\noindent \textbf{Generalizable Bimanual Manipulation.}
Bimanual manipulation agents~\citep{bersch2011bimanual, Kataoka2022bimanual, fu2024mobilealohalearningbimanual, zhang2021dairdisentangledattentionintrinsic, yu2024bikc, gbagbe2024bivlavisionlanguageactionmodelbasedbimanual, liu2024voxactbvoxelbasedactingstabilizing, gao2024bikvilkeypointsbasedvisualimitation,grotz2024peract2benchmarkinglearningrobotic, chen2023bi} are able to handle a large variety of tasks by predicting a trajectory of bimanual operation, which is of great significance in complex applications from household service \citep{zhang2024empoweringembodiedmanipulationbimanualmobile}, robotic surgery \citep{10149474}, to component assembly in factories \citep{buhl2019dual}.
To achieve multi-task learning for generalizable Bimanual manipulation, earlier studies attempted to leverage the emerged general understanding and reasoning capacities of pretrained foundation models like LLMs \citep{touvron2023llama2openfoundation} and VLMs \citep{chen2023pali3visionlanguagemodels}, where the foundation model was prompted to generate a high-level plan for low-level executors. 
% For example, VoxPoser \citep{huang2023voxposercomposable3dvalue} utilized LLMs and VLMs to specify 3D affordance and constraint map via code generation, where a sampling-based motion planner is employed as the low-level executor. 
However, the performance of directly leveraging foundation models in a training-free manner is bottlenecked by the predefined low-level executor, which struggles to generalize to more contact-rich tasks like straightening a rope.
% that are highly-desired in real-world applications.
To overcome this challenge, robotic foundation models \citep{kim2024openvlaopensourcevisionlanguageactionmodel, liu2024rdt, embodimentcollaboration2024openxembodimentroboticlearning, rt22023arxiv, brohan2022rt, octomodelteam2024octoopensourcegeneralistrobot} that pretrained on large-scale real-world demonstrations were proposed under the unimanual setting, which has shown high generalizability across everyday manipulation tasks.
% earlier studies~\citep{figueroa2017learning, ureche2018constraints, batinica2017compliant, colome2018dimensionality, colome2020reinforcement, franzese2022interactive, chitnis2020efficient} explored reinforcement learning (RL) for these tasks, which needs labor-intensive manual-crafted reward functions and often takes enormous training time to learn a useful policy. Consequently, recent works [xx] started to pay attention to learning from expert demonstrations with imitaion learning (IL).
%Compared with unimanual settings, the crux of learning bimanual manipulation is the fine-grained control of high-dimensional robot action, which makes teleoperating bimanual system to collect demonstrations for generalizable policy training extremely expensive.
% Unlike unimanual manipulation
However, bimanual tasks demand precise coordination of two high degree-of-freedom arms, making the teleoperation of demonstrations for training generalizable policies also costly.
% Hard-ware
Although some recent approaches~\citep{tony2023aloha, fu2024mobilealohalearningbimanual, ding2024bunnyvisionprorealtimebimanualdexterous, chuang2024activevisionneedexploring, yang2024acecrossplatformvisualexoskeletonslowcost, chi2024universal} have developed more specialized teleoperation systems to reduce these costs, scaling up demonstrations for high generalization ability remains a challenge.
% To directly reduce the cost of teleoperating two high-degree-freedom arms, Some methods ~\citep{tony2023aloha}[mobile aloha] designed more efficient and human-friendly hardwares, but the expert demonstrations is still hard to scale up with specialized systems.
% stablizing and acting
To address the limited availability of demonstrations, alternative methods~\citep{wang2021influencing, liu2024voxactbvoxelbasedactingstabilizing, grannen2023stabilizeactlearningcoordinate} proposed to 
simplify the learning of bimanual policies by decoupling the bimanual system into a stabilizing arm and an acting arm.
Nevertheless, these methods often rely on predefined roles for each arm, which precludes their applicability to tasks requiring more flexible collaboration patterns.
In contrast to these approaches, our work presents a novel method that transfers generalizable unimanual policies to bimanual tasks, which eliminates the necessity for explicit inductive bias like role specification.
% This strategy reduces the need for extensive bimanual demonstrations and eliminates the necessity for explicit inductive bias like role specification, thereby providing a more adaptable solution for complex tasks.

% ~\citep{figueroa2017learning, ureche2018constraints, Zollner2004programming, smith2012dual, stepputtis2022system, lioutikov2016learning}. 

% learning RL from scratch
% RL methods for learning high-frequency bimanual control policies can require a large number of samples and many hours of robot time, which makes simulation to real policy transfer an appealing approach~\citep{chitnis2020efficient, Kataoka2022bimanual, chen2022towards}. However, sim-to-real approaches are limited to settings where the sim-to-real gap is small, which precludes many contact-rich bimanual tasks such as zipping a zipper or cutting food~\citep{stepputtis2022system, elguea2023review, kroemer2021review}.
% shrink the difficulty of learning high-dimensional bimanual policy by decoupleing the bimanual system to a stabilizing arm and an acting arm.

\begin{figure*}[t]
    \centering
    \includegraphics[width=1\textwidth]{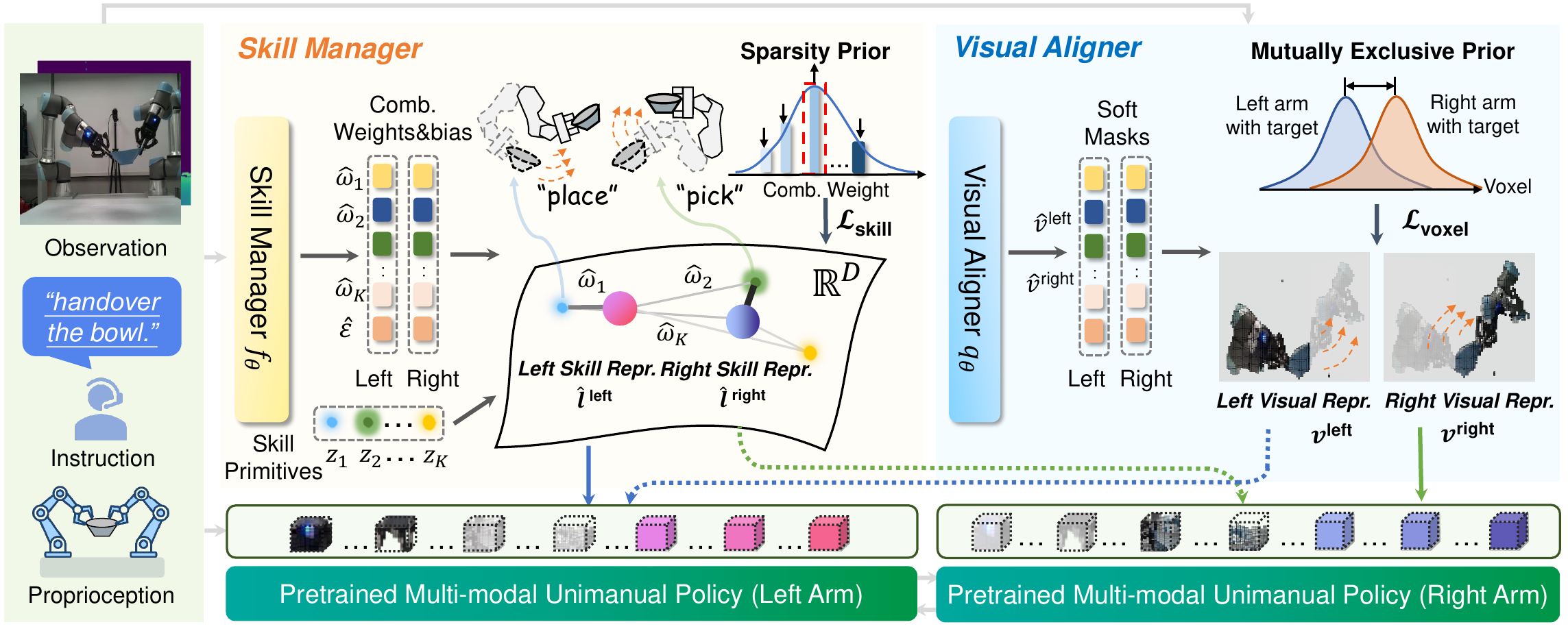}
    \caption{\small \textbf{The overall pipeline of \method}, which primarily consists of a skill manager and a perception manager. The skill manager adaptively coordinates primitive skills for each robot arm, while the perception manager mitigates the distributional shift from unimanual to bimanual by decomposing the 3D voxel observation for each arm.
    }
    \label{fig:pipeline}
\end{figure*}

\noindent \textbf{Skill-based Methods.}
Skill learning \citep{zhang2023bootstrapskillslearningsolve} is the process where intelligent agents acquire new abilities that are transferable across different tasks, which is of great significance for cross-task generalization. Thus, skill learning is being attractive in enhancing the generalizability of different models, such as game agents \citep{wang2023voyageropenendedembodiedagent}, robotic manipulation \citep{liang2024skilldiffuserinterpretablehierarchicalplanning}, and autonomous driving \citep{fu2023drivelikehumanrethinking}.
The initial attempt to utilize skill learning was orchestrating a set of predefined skill \citep{munawar2018maestrob}, which hindered their scalability to unseen tasks.
To overcome this limitation, \citep{liang2024skilldiffuserinterpretablehierarchicalplanning, chitnis2020efficient} proposed to learn shareable skill primitives from data.
For example, skill diffuser \citep{liang2024skilldiffuserinterpretablehierarchicalplanning} introduced a hierarchical planning framework that integrates learnable skill embedding into conditional trajectory generation, which realized accurate execution of diverse compositional tasks.
% In the domain of skill learning through language instructions, LISA [13] stands out by sampling multiple skills per trajectory, integrating language conditioning in a unique manner.
In the field of bimanual manipulation, skill learning was dominated by handcrafted primitives. For example, \citep{batinica2017compliant, franzese2022interactive, grannen2020untangling, avigal2022speedfolding, ganapathi2020learning, amadio2019exploiting, yin2014learning, fu2022safely, Chitnis2020Intrinsic, xie2020deep, ha2021flingbot} propose to utilize parameterized atomic movements to shrink the high dimensionality of the bimanual action space, which has shown impressive performance on templated bimanual manipulation tasks. 
While the predefined atomic movements did boost the success rate on specific tasks, they are often difficult for even human users to specify, which restricts the deployment scenarios of these methods.
% they limits the generalizability to complex tasks that can not be easily represented by explicit movements.
In this paper, we propose leveraging learnable skill primitives to represent the learned commonsense of the pretrained unimanual policy, so that the knowledge can be transferred across different levels of tasks.

% What is skill learning? application?

% Early effort

% To address these limitations

% In contrast to 

% \noindent \textbf{Parameter-Efficient  Fine-Tuning.}
% \noindent \textbf{Visual Decomposition Methods.}

% ??

% Take a deep breath, I want you to act as an experienced academician. I'm writing a scientific paper for CVPR conference. Here is the approach section of my draft, please revise it to make it more fluent and compatible with CVPR conference. Only use simple words and expressions. This is very important to my career. """ """

\section{Approach}
\label{sec:approach}

In this section, we first introduce required preliminaries (\Cref{subsec:prelimilaries}), and then we present a pipeline overview (\Cref{subsec:overall}).
Then, we introduce a skill manager (\Cref{subsec:skill_manager}) that schedules skills to each arm to form general bimanual manipulation policies. We build a visual aligner (\Cref{subsec:perception_manager}) to mitigate the observation discrepancy between bimanual and unimanual systems for better generalization.
Finally, we outline the training objectives (\Cref{subsec:learning_objectives}).

% policy generalization ability enhancement.

\subsection{Problem Formulation}\label{subsec:prelimilaries}

%Language-conditioned bimanual manipulation is of great significance in the pursuit of the general intelligent humanoid robot. 
The task of policy learning for bimanual manipulation can be defined as follows. To complete a wide range of manipulation tasks specified in natural language, the bimanual agent is required to interactively predict the actions of both end-effectors based on the visual observation and robot states, where the motion is acquired by a low-level planner (\eg, RRT-Connect). The observation $o_t$ at the $t_{th}$ time step includes the voxel $v_t$ converted from RGB and depth images \citep{grotz2024peract2benchmarkinglearningrobotic, shridhar2022perceiveractormultitasktransformerrobotic} and the robot proprioception $p_t$. 
The action $a_t$ for each end-effector at the $t_{th}$ time step contains the location $a_{\text{trans}}$, orientation $a_{\text{rot}}$, gripper open state $a_{\text{open}}$ and usage of collision avoidance in the motion planner $a_{\text{col}}$.
% For the training data, human demonstrators produce $K$ offline expert trajectories $\mathcal{D} = \{(o_1, a^{\text{left}}_1, a^{\text{right}}_1), ..., (o_K, a^{\text{left}}_K, a^{\text{right}}_K)\}$ for each task instruction, where $a^{\text{left}}_t$ and $a^{\text{right}}_t$ respectively demonstrate the actions for left and right grippers. 
For the training data, human demonstrators produce a limited set of $M$ offline expert trajectories $\mathcal{D} = \{(o_1, a^{\text{left}}_1, a^{\text{right}}_1), ..., (o_M, a^{\text{left}}_M, a^{\text{right}}_M)\}$ for each task instruction $l$, where $a^{\text{arm}}_t, \text{arm}\in \mathcal{A}=\{\text{left}, \text{right}\}$ demonstrates the actions for left and right grippers.
% Each trajectory also pairs with a natural language instruction $l$ that specifies the goal.
Existing methods directly learn the policy model from expert demonstrations, which have shown effectiveness in single-task settings.
However, due to the high cost of data collection in bimanual systems, the scarcity of expert demonstrations limits the generalizability of these methods across tasks.
To address this, we present to transfer pretrained generalizable unimanual policy for general bimanual manipulation.
%a novel plug-and-play method named AnyBimanual, which enables transferring from pretrained unimanual policies to generalizable multi-task bimanual manipulation policies.

% Many methods have been proposed to learn such an agent, including imitation learning and reinforcement learning. 
% Each action $a_t$ is composed of a left arm pose $a^{\text{left}}_t$ and a right arm pose $a^{\text{right}}_t$.

\subsection{Overall Pipeline}\label{subsec:overall}

Overall pipeline of our \method method is shown in \Cref{fig:pipeline}. For the language branch, we employed a pretrained text encoder \citep{radford2021learningtransferablevisualmodels} to parse the bimanual instruction to language embeddings with high-level semantics, where the skill manager schedules the skill primitives with composition and compensation to enhance the language embeddings that instruct relevant subtasks for different arms. Therefore, the pre-trained unimanual policy model can be prompted to generate feasible manipulation policy for each arm with high generalization ability across tasks with the sharable manipulation knowledge. For the visual branch, we voxelize the RGB-D input to the voxel space as observation, and a 3D sparse voxel encoder is utilized to tokenize the voxel observation for informative volumetric representation. 
The visual aligner generates a soft spatial mask to align the visual representation of unimanual policy model with its representation during pretraining, so that the observation discrepancy between unimanual and bimanual systems can be minimized for policy transferability enhancement. 
We employ two pretrained unimanual models to predict the left and right robot actions based on the text embeddings and visual representations, where the pretrained unimanual policy can be multi-modal transformer-based policy \citep{shridhar2022perceiveractormultitasktransformerrobotic, kim2024openvlaopensourcevisionlanguageactionmodel, rt22023arxiv, brohan2022rt, embodimentcollaboration2024openxembodimentroboticlearning} or diffusion-based policy \citep{ke20243ddiffuseractorpolicy, octomodelteam2024octoopensourcegeneralistrobot}.

% We present two model-agnostic adapter called skill manager and perception manager to 

% \subsection{Discovering and Scheduling Transferable unimanual Skill Primitives}\label{subsec:skill_manager}
\subsection{Scheduling Unimanual Skill Primitives}\label{subsec:skill_manager}

In order to transfer unimanual manipulation policy to bimanual settings without generalizability drops, we propose a skill manager to decompose the action policies from unimanual foundation models into skill primitives and integrate primitives for bimanual systems. 
However, the given offline expert demonstrations $\mathcal{D}$ do not contain any explicit intermediate skill primitives or sub-task boundaries, but only low-level end-effector poses and high-level natural language instruction are provided. Therefore, we design an automatic skill discovery method in an unsupervised manner to learn skill representations and their schema from offline bimanual manipulation datasets during training. In the test phase, the skill manager predicts different weighted combinations of primitive skills to orchestrate each arm given high-level language instruction, which enables effective transfer of pre-trained unimanual policy to diverse bimanual manipulation tasks.

% \noindent\textbf{Skill Discovery}
%\noindent\textbf{Skill Representation Theroem}
\noindent\textbf{Skill Manager.}
We start with a discrete primitive skill set $\mathcal{Z} = \{z_1, z_2, ..., z_K\}$, where $K$ is a hyper-parameter that indicates the number of skill primitives. 
% A large $K$ may improve the expression ability of skill subspace, while a small $K$ may enhance the sparsity of the skill representation. 
% The trade-off between the expression ability of skill subspaces and the sparsity of the skill representation is considered by tuning the optimal $K$.
To realize end-to-end skill discovery and scheduling, each potential skill is an implicit embedding $z_k \in \mathbb{R}^{D}$, 
% which can be either randomly initialized or initialized with the corresponding language template tokens of the pretrained unimanual policy to mitigate the domain gap.
which can be initialized with the corresponding language template tokens of the pretrained unimanual policy to mitigate the domain gap.
\begin{figure}[t]
% \vspace{-0.3cm}
    \centering
    \includegraphics[width=0.48\textwidth]{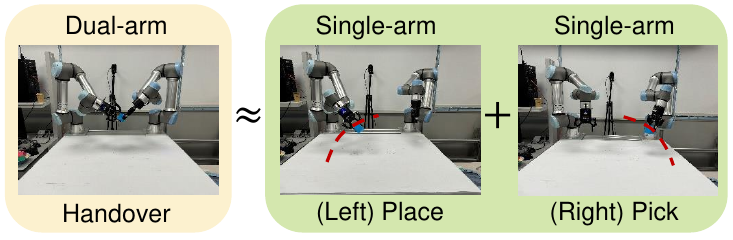}
    % \vspace{-0.5cm}
    \caption{\small \textbf{Shareable skills across unimanual and bimanual settings.} We observe that bimanual tasks are often originated from the combination of unimanual sub-tasks, which thus can be solved by effectively coordinating unimanual skills synchronously or asynchronously.}
    \label{fig:toy_example}
    \vspace{-0.4cm}
\end{figure}
By combining the primitives from the skill set, the language embedding for the unimanual policy model can be represented as a linear combination of these primitives. 
% For both values of $\text{arm} \in \{\text{left}, \text{right}\}$,
Hence, the reconstructed language embedding as skill representation can be expressed as:
% \begin{equation}
% \hat{l}^{\text{arm}}_t = \sum^K_{k=1} \hat{w}^{\text{arm}}_{k,t} z_k + \epsilon^{\text{arm}}_t,
% \end{equation}
\begin{equation}
\hat{l}^{\text{left}}_t = \sum^K_{k=1} \hat{w}^{\text{left}}_{k,t} z_k + \epsilon^{\text{left}}_t, \quad
\hat{l}^{\text{right}}_t = \sum^K_{k=1} \hat{w}^{\text{right}}_{k,t} z_k + \epsilon^{\text{right}}_t
\end{equation}
where $\hat{l}^{\text{arm}}_t$ is the decomposed unimanual language embedding for one arm of the bimanual system, and $\hat{w}^{\text{arm}}_t \in \mathbb{R}^{K}$ denotes the importance weight for linear combination
for both values of $\text{arm} \in \mathcal{A}$. We also introduce a task-oriented compensation  $\epsilon^{\text{arm}}_t \in \mathbb{R}^D$ to introduce the embodiment-specific knowledge for the policy transfer. The upscript arm of the variables can be selected from left or right to indicate the embodiment in the bimanual systems. 
% For example, the compensation term works for asymmetrical tasks like \texttt{sweep dustpan} where an arm should act as a `stablizer' [stab and act] to stablize the dustpan while another arm is sweeping the rubbish rarely occurs in unimanual settings.
%% Skill representation theroem
As depicted in \Cref{fig:toy_example}, considering a bimanual task \texttt{Handover}, it can be explicitly solved by scheduling two unimanual primitive skills, i.e., the left arm \texttt{Place} the block to the right gripper while the right arm \texttt{Pick} it from the left gripper. 

%\noindent\textbf{Skill Manager}
% $\hat{l}^{\text{arm}}_t$
Though every language embedding that passed through the pretrained single policy can be represented as a linear combination of the skill set, the combination weights that specify the importance of each skill are dynamic across the task completion process. We propose to parametrize a multi-model transformer named skill manager to dynamically predict the combination weight for each arm at each time step.
Therefore, our skill manager $f$ can be formulated as $(\hat{w}^{\text{left}}_t, \epsilon^{\text{left}}_t, \hat{w}^{\text{right}}_t, \epsilon^{\text{right}}_t) = f_\theta(v_t, l, p_t)$, which takes the overall bimanual visual and language embeddings, proprioception as input, and assigns the reconstructed unimanual language embedding for each arm as output to schedule the skill primitive of each arm dynamically. $\theta$ represents the learnable parameters.
Finally, the combined skill primitives are concatenated with the initial bimanual language embedding to enhance the global context, which is then forwarded to the corresponding unimanual policy.
% Both the predicted skill embedding and the original language embedding are then con forwarded to the unimanual model. 
% i.e., $l^{\text{arm}}_t = \hat{l}^{\text{arm}}_t \bigoplus l$, where $\bigoplus$ refers to the concatenation operator.

% \noindent\textbf{Learning Generalizable Skill Primitives.}
\noindent\textbf{Learning Generalizable Skill Representations.}
 % via Sparse Representation
% without redundance
% The skill manager could be trained with the overall behavior cloning loss. 
To update the skill library, we expect the discorverd skill representations are informative to encode fundamental robot motions that are sharable across a variety of tasks, thereby enhancing the generalizability of our framework.
To realize this, the learning objective of the skill manager is designed as a sparse representation problem~\citep{wright2010sparse}:
% \begin{equation}
%     \mathcal{L}_{\text {skill}} = \| \hat{w}^{\text{arm}} \|_{1} + \lambda_{\epsilon} \| \epsilon^{\text{arm}} \|_{2,1}
% \end{equation}
\begin{equation}
    \mathcal{L}_{\text {skill}} = \| \hat{w}^{\text{left}} \|_{1} + \| \hat{w}^{\text{right}} \|_{1} + \lambda_{\epsilon} (\| \epsilon^{\text{left}} \|_{2,1} + \| \epsilon^{\text{right}} \|_{2,1})
\end{equation}
where $\|\cdot\|_1$, and $\|\cdot\|_{2,1}$ denote the $\ell_1$ and $\ell_{2,1}$ norm, respectively. $\lambda_{\epsilon}$ is a hyper-parameter that balances the denoising term.
%% plain explain 
By minimizing this sparse regularization term, the skill manager is encouraged to reconstruct the skill representation by selecting fewer skill primitives, which is further surrogated by minimizing a differentiable $\ell_1$ regularization \citep{tibshirani1996regression}.
%% logic: sparse -> informative
Therefore, the skill subspaces are required to be orthogonal and disjoint with each other to reconstruct the language embedding with sparse combination and compensation, which implicitly enforces each skill representation to capture an independent fundamental motion.
% To encourage the orthogonality of the learned skill primitives, we present to learn a sparse representation for the language embedding, which means the disjoint subspace.
% Finally, the agent is encouraged to mine informative and discriminative skill basis by minimizing this sparse regularization term, which are sharable across a variety of tasks.

% \subsection{Decomposing Volumetric Observation with Spatial Soft Mask}\label{subsec:perception_manager}
\subsection{Aligning Unimanual Visual Representations}\label{subsec:perception_manager}

\noindent\textbf{Visual Aligner.}
Despite the skill manager enabling generalization in the language modality, the distributional shift from the unimanual to bimanual workspace in terms of the visual context still may harm the model performance.
To mitigate the observation discrepancy, we present a visual aligner $q$ that predict two spatial soft masks at each step $t$ to edit the voxel space so that the decomposed subspace of unimanual policy model for each arm aligns with those during pretraining stage: $(\hat{v}^{\text{left}}_t, \hat{v}^{\text{right}}_t) = q_\theta(v_t, l, p_t)$.
The decomposed observation represents the locality of the workspace, which is then augmented by the bimanual observation to form the final visual embedding: 
% \begin{equation}
% v^{\text{arm}}_t = (\hat{v}^{\text{arm}}_t \odot v_t) \oplus v_t, 
% \end{equation}
\begin{equation}
v^{\text{left}}_t = (\hat{v}^{\text{left}}_t \odot v_t) \oplus v_t, \quad v^{\text{right}}_t = (\hat{v}^{\text{right}}_t \odot v_t) \oplus v_t, 
\end{equation}
where $\odot$ is the element-wise multiplication, and $\oplus$ refers to the concatenation operator. As a result, the augmented visual representations for each arm contains both embodiment-specific information and the global context, which is then passed through the two unimanual policy models to decode the optimal bimanual action. 

% \noindent\textbf{Learning Aligned Spatial Representation.}
\noindent\textbf{Learning Aligned Visual Representations.}
Our goal is to mitigate the visual domain gap between the unimanual and bimanual setting, so that the pretrained commonsense knowledge in unimanual policy can be transferred with high adaptation ability.
%% high-level idea
Since we can not access the unimanual pretraining data in common usages, we instead impose a mutually exclusive prior to the visual aligner.
% encourages it to implicitly divide the bimanual workspace into unimanual workspaces.
This prior is regularized by optimizing a Jensen-Shannon (JS) divergence regularization term:
\begin{equation}
    \mathcal{L}_{\text {voxel}} = - D_{KL}(\hat{v}^{\text{left}}_t \| \hat{v}^{\text{right}}_t)/2 - D_{KL}(\hat{v}^{\text{right}}_t \| \hat{v}^{\text{left}}_t)/2
\end{equation}
where $D_{KL}$ means the Kullback-Leibler (KL) divergence operator.
% Bimanual manipulation tasks often involves unsynchronized collaboration that requires the left and right arm attends to different areas of the whole workspace to act as different roles, such as stabilizing and acting. 
% To account for this, we encourage the visual aligner to assign complementary volumetric representation by including a Kullback-Leibler (KL) divergence or M-projection regularization term:
%% Focus on different area = left arm, right arm
To provide further explanation, bimanual manipulation tasks often involve asynchronous collaboration that requires the left and right arm to attend to different areas of the whole workspace to act as different roles, such as stabilizing and acting. 
As a result, the mutually exclusive division of the entire bimanual workspace will naturally separate one arm and its target from the other, which highly resembles the unimanual configuration.
% Hence by maximizing the divergence between the two soft masks that represents the volumetric importance for each arm, the voxel input of the bimanual manipulation agent can be disentangled into unimanual visual representations that align with those in the pretraining phase effectively.
Hence by maximizing the divergence between the two soft masks, the voxel input of the bimanual manipulation agent can be disentangled into unimanual visual representations that align with those in the pretraining phase effectively.

\subsection{Learning Objectives}\label{subsec:learning_objectives}
The decomposed skill and volumetric representation are used to pass through the two pretrained unimanual policies to predict the optimal actions of the two end-effectors.
% \noindent\textbf{Orchestrating unimanual Policies}
We assume access to a pretrained unimanual policy $p$, which is fundamentally a multi-model multi-task neural network that takes visual and language embedding as inputs and outputs end-effector actions.
% $a^{\text{arm}}_t = p(v^{\text{arm}}_t, l^{\text{arm}}_t, p^{\text{arm}}_t), \text{arm} \in \{\text{left}, \text{right}\}$.
Our \method is a model-agnostic plug-and-play method, which indicates that the architecture of pretrained unimanual policy $p$ is flexible in different architectures such as multi-modal transformer-based policies~\citep{shridhar2022perceiveractormultitasktransformerrobotic, kim2024openvlaopensourcevisionlanguageactionmodel} and diffusion policies \citep{ke20243ddiffuseractorpolicy, octomodelteam2024octoopensourcegeneralistrobot}.
To supervise the model with the provided expert demonstrations for behavior cloning, we leverage the cross-entropy loss to learn accurate action prediction for each arm:
% \begin{equation}
% \small
% \mathcal{L}_{\text {BC}}\!=\!CE(p^{\text{left}}_{\text{trans}}, p^{\text{left}}_{\text{rot}}, p^{\text{left}}_{\text{open}}, p^{\text{left}}_{\text{col}}) \!+\! CE(p^{\text{right}}_{\text{trans}}, p^{\text{right}}_{\text{rot}}\!, p^{\text{right}}_{\text{open}}, p^{\text{right}}_{\text{col}})
% \end{equation}
\begin{equation}
\mathcal{L}_{\text {BC}} = \sum_{\text{arm}\in\mathcal{A}} CE(p^{\text{arm}}_{\text{trans}}, p^{\text{arm}}_{\text{rot}}, p^{\text{arm}}_{\text{open}}, p^{\text{arm}}_{\text{col}})
\end{equation}
where $p
^{\text{arm}}_{\text{trans}}, p^{\text{arm}}_{\text{rot}}, p^{\text{arm}}_{\text{open}}, p^{\text{arm}}_{\text{col}}$ represents the distribution of the ground-truth actions for translation, rotation, gripper openness, and collision avoidance for the corresponding robot arm, respectively. The behavior cloning loss is then combined with the two regularization terms described above to learn informative skill manager and visual aligner.
To sum up, the training objective of \method is:
\begin{equation}
\mathcal{L}_{\text {total}} = \mathcal{L}_{\text {BC}} + \lambda_{\text{skill}} \mathcal{L}_{\text {skill}} + \lambda_{\text{voxel}} \mathcal{L}_{\text {voxel}},
\end{equation}
where $\lambda_{\text{skill}}$ and $\lambda_{\text{voxel}}$ refer to hyper-parameters that balance the importance of the regularizations.

% Take a deep breath, I want you to act as an experienced academician. I'm writing a scientific paper for CVPR conference. Here is the experiment of my draft, please revise it to make it more fluent and compatible with CVPR conference. Only use simple words and expressions. This is very important to my career. """ """
% \newcommand{\tabref}[1]{Table~\ref{#1}}

\begin{table*}[t]
% \vspace{-1.0cm}
\centering
\scriptsize
% \small
\caption{\small  \textbf{Multi-Task Test Results in Simulator.} Average success rates (\%) of general bimanual manipulation agents trained with $20$ or $100$ demonstrations per task and evaluated over $100$ episodes. `LF' means the leader-follower architecture that transfers unimanual manipulation policy for bimanual manipulation. 
% The average performances are shown in \Cref{fig:teaser}.
% Each evaluation is scored as either 0 for failure or 100 for success.
\method enables plug-and-play transfers for multiple unimanual manipulation policies.
}
\setlength{\tabcolsep}{12pt} % scriptsize
% \setlength{\tabcolsep}{9pt}
% \renewcommand{\arraystretch}{0.9} 
% \resizebox{\linewidth}{!}{
\begin{tabular}{lcccccccccccc} 
% \toprule
\shline
                  & \multicolumn{2}{c}{\begin{tabular}[c]{@{}c@{}}\texttt{pick} \\\texttt{laptop}\end{tabular}}     & \multicolumn{2}{c}{\begin{tabular}[c]{@{}c@{}}\texttt{pick} \\\texttt{plate}\end{tabular}} &  \multicolumn{2}{c}{\begin{tabular}[c]{@{}c@{}}\texttt{straighten} \\\texttt{rope}\end{tabular}}   &
                  \multicolumn{2}{c}{\begin{tabular}[c]{@{}c@{}}\texttt{lift}\\\texttt{ball}\end{tabular}} &
                  \multicolumn{2}{c}{\begin{tabular}[c]{@{}c@{}}\texttt{lift} \\\texttt{tray}\end{tabular}}   & 
                  \multicolumn{2}{c}{\begin{tabular}[c]{@{}c@{}}\texttt{push} \\\texttt{box}\end{tabular}} 
                         
                  \\
                  \cmidrule(lr){2-3} \cmidrule(lr){4-5} \cmidrule(lr){6-7} \cmidrule(lr){8-9} \cmidrule(lr){10-11} \cmidrule(lr){12-13} 
                  
                  \\[-13pt]                                                         \\
\vcell{Method}  & \vcell{20}  & \vcell{100}  & \vcell{20}  & \vcell{100}  & \vcell{20}  & \vcell{100}  & \vcell{20}  & \vcell{100}  & \vcell{20}  & \vcell{100}  & \vcell{20}  & \vcell{100} \\[-\rowheight] 
\printcellbottom  & \printcellbottom & \printcellbottom  & \printcellbottom & \printcellbottom  & \printcellbottom & \printcellbottom   & \printcellbottom & \printcellbottom   & \printcellbottom & \printcellbottom   & \printcellbottom  & \printcellbottom  \\[0pt] 
\hline \\[-6pt]
% \shline \\[-6pt]
RVT~\citep{goyal2023rvtroboticviewtransformer}-LF & 1  & 2 & 1  & 1 & 1 & 2 & 3 & 3 & 4 & 6 & 7 & 18 \\
\rowcolor{\ourcolor}
\textbf{RVT-LF + \method}  & 1  & 3 &  2 & 4 & 2 & 5 & 3 & 4 & 4 & 6 & 11 & 21 \\
\hline \\[-6pt]
PerAct~\citep{shridhar2022perceiveractormultitasktransformerrobotic}-LF & 1  & 2 &  3 & 4 & 5 & 11 & 4 & 7 & 6 & 12 & 7 & 17 \\
\rowcolor{\ourcolor}
\textbf{PerAct-LF + \method} & 2 & 2 & 3 & 5 & 12 & 14 & 8 & 8 & \textbf{15} & \underline{17} & \underline{23} & 29  \\
\hline \\[-6pt]
PerAct$^2$~\citep{grotz2024peract2benchmarkinglearningrobotic} & \underline{3} & 4 & 2 & 4 & 6 & 8 & 4 & 4 & 4 & 3 & 5 & 6  \\
PerAct$^2$ + Pretraining & \underline{3} & \underline{5} & \underline{5} & \underline{7} & \underline{13} & \underline{17} & \underline{9} & \underline{14} & 2 & 5 & \underline{23} & \underline{31} \\
\rowcolor{\ourcolor}
\textbf{PerAct + \method} & \textbf{4} & \textbf{7} & \textbf{6} & \textbf{8} & \textbf{17} & \textbf{24} & \textbf{22} & \textbf{36} & \underline{9} & \underline{14} & \textbf{31} & \textbf{46} \\[1pt]
\hline \\[-6pt]
% \shline \\[-6pt]
                  & \multicolumn{2}{c}{\begin{tabular}[c]{@{}c@{}}\texttt{put in} \\\texttt{fridge}\end{tabular}} &
                  \multicolumn{2}{c}{\begin{tabular}[c]{@{}c@{}}\texttt{press}\\\texttt{buttons}\end{tabular}} & \multicolumn{2}{c}{\begin{tabular}[c]{@{}c@{}}\texttt{handover}\\\texttt{item}\end{tabular}}      & \multicolumn{2}{c}{\begin{tabular}[c]{@{}c@{}}\texttt{sweep to}\\\texttt{dustpan}\end{tabular}} & 
                  \multicolumn{2}{c}{\begin{tabular}[c]{@{}c@{}}\texttt{take out}\\\texttt{tray}\end{tabular}} &
                  \multicolumn{2}{c}{\begin{tabular}[c]{@{}c@{}}\texttt{handover}\\\texttt{easy}\end{tabular}}
                \\
                  \cmidrule(lr){2-3} \cmidrule(lr){4-5} \cmidrule(lr){6-7} \cmidrule(lr){8-9} \cmidrule(lr){10-11} \cmidrule(lr){12-13} 
                  \\[-6pt]          
\vcell{} & \vcell{20} & \vcell{100} & \vcell{20} & \vcell{100} & \vcell{20} & \vcell{100} & \vcell{20} & \vcell{100} & \vcell{20} & \vcell{100} & \vcell{20} & \vcell{100}  \\[-\rowheight]
\printcellbottom & \printcellbottom & \printcellbottom & \printcellbottom & \printcellbottom & \printcellbottom & \printcellbottom & \printcellbottom & \printcellbottom & \printcellbottom & \printcellbottom & \printcellbottom & \printcellbottom \\[1pt]
\hline \\[-6pt]
% \shline \\[-6pt]
RVT~\citep{goyal2023rvtroboticviewtransformer}-LF & 0  & 0 & 10  & 22 & 0 & 0 & 0 & 0 & 2 & 2 & 3 & 3 \\
\rowcolor{\ourcolor}
\textbf{RVT-LF + \method} & 1  & 3 &  17 & 32 & 1 & 2 & 4 & 13 & 1 & 2 & 2 & 7 \\
\hline \\[-6pt]
PerAct~\citep{shridhar2022perceiveractormultitasktransformerrobotic}-LF & 2 & 7 & 2 & 9 & 0 & 3 & 22 & 47 & 0 & 3 & 2 & 5 \\
\rowcolor{\ourcolor}
\textbf{PerAct-LF + \method} & 4 & 9 & 12 & 14 & 1 & 5 & \underline{34} & \underline{57} & 1 & 4 & 3 & 13 \\
\hline \\[-6pt]
PerAct$^2$~\citep{grotz2024peract2benchmarkinglearningrobotic} & 7 & 16 & 23 & 41 & \underline{3} & 6 & 25 & 52 & 1 & 3 & 22 & 29 \\
PerAct$^2$ + Pretraining & \underline{10} & \underline{22} & \underline{27} & \underline{40} & \underline{3} & \underline{7} & 29 & 55 & \underline{3} & \underline{8} & \underline{24} & \underline{33}  \\
% \rowcolor[rgb]{0.9,1.0,0.9}
\rowcolor{\ourcolor}
\textbf{PerAct + \method} & \textbf{13} & \textbf{26} & \textbf{39} & \textbf{73} & \textbf{7} & \textbf{15} & \textbf{43} & \textbf{67} & \textbf{9} & \textbf{24} & \textbf{31} & \textbf{44}  \\
\bottomrule
% \shline
%%%%%% \textbf{PerAct$^2$ + \method}
\end{tabular}
\vspace{-0.1cm}
\label{table:results}
\end{table*}

\begin{table}[t]
  \centering
  \scriptsize
  % \small
    % \caption{\small \textbf{Comparison of Our Methods with Different Techniques.} 
    \caption{\small \textbf{Comparison of \method with Different Techniques.} 
    % \caption{\small \textbf{Ablation Study on Different Techniques.} 
    % We manually categorize the $12$ RLBench2 task to $6$ groups for further interpretability. 
    We categorize the long-term tasks that require more than \num{6.5} keyframes to \texttt{Long}, tasks that involve multiple variations to \texttt{Generalized} and tasks that involve synchronization of both arms to \texttt{Sync} for further interpretability.
    % For more details regarding the classification method, please refer to the supplementary file.
    }
    \vspace{-0.2cm}
  \setlength{\tabcolsep}{1.7pt} 
  % \resizebox{0.49\textwidth}{!}{ % Scale the table to fit within text width
     \begin{tabular}{c|cc|ccc|c}
    \toprule
    Row ID & Skill Manager & Visual Aligner & \texttt{Long} & \texttt{Generalized} & \texttt{Sync} & \textbf{Average} \\
    \midrule
    1 & - & - & 16.29 & 23.50 & 3.50 &  14.67 \\
    \hline
    2 & \no & \no & 19.57 & 25.50 & 9.50 &  16.75 \\
    3 & \no & \yes & 21.57 & 44.00 & 15.50 &  19.75 \\
    4 & \yes & \no & 23.71 & 42.00 & 17.00 &  \underline{25.67} \\
    \rowcolor{\ourcolor}
    5 & \yes & \yes & 27.29 & 44.00 & 25.00 &  \textbf{32.00} \\
    \bottomrule
    \end{tabular}
  % }
    \vspace{-0.3cm}
  \label{table:ablation}
\end{table}

\section{Experiments}
\label{sec:experiments}

% including datasets, baseline methods and implementation details 
In this section, we first introduce the experimental setups (\Cref{subsec:setup}). Then, we transfer various unimanual methods to bimanual manipulation via \method, and compare them with the state-of-the-art to show uperiority (\Cref{subsec:comp_with_sota}). 
We conduct an ablation study to evaluate validity of the proposed components (\Cref{subsec: ablation_study}). We further interpret learned skill representation and decomposed volumetric representation by visualization (\Cref{subsec:qualitative}). 
Finally, we report real-robot results to demonstrate effectiveness of \method in real-world applications (\Cref{subsec:real_robot_results}).

% More results and case studies can be found in the supplementary file.

\subsection{Experiment Setup}\label{subsec:setup}
\textbf{Simulation.} For benchmarking, our simulation experiments are conducted on  RLBench2~\citep{grotz2024peract2benchmarkinglearningrobotic}, a bimanual version extended from the widely-used RLBench \citep{james2019rlbenchrobotlearningbenchmark} benchmark in prior works~\citep{jaegle2021perceivergeneralperceptioniterative, ze2024gnfactormultitaskrealrobot, lu2024manigaussiandynamicgaussiansplatting, goyal2023rvtroboticviewtransformer, xian2023chaineddiffuser, ke20243ddiffuseractorpolicy}. Following the setup in \citep{grotz2024peract2benchmarkinglearningrobotic}, we utilize $12$ language-conditioned bimanual manipulation tasks varying from different challenge levels. 
The diverse task suite requires the agent to acquire and correctly schedule shareable skills to achieve high success rates, rather than merely imitating limited expert demonstrations. 
% based on natural language instructions
% (\ie, front, left, right, wrist left, wrist right, overhead)
For observation, we employ six cameras with a resolution of $256 \times 256$ to cover the entire workspace. 
During the training phase, we provide $20$ or $100$ demonstrations for each task, and we evaluate $100$ episodes per task in the testing set to mitigate the randomness.
% , which are generated by an oracle script. 
% In the test phase, we evaluate $25$ episodes per task in the testing set to mitigate bias from noise.

% as 6-DoF controllers
% mounted on a tripod
% simulating a human viewpoint during bimanual tasks
% Camera-to-arm calibration is achieved using the easy\_handeye package and ARUCO markers on the end-effectors.
\noindent\textbf{Real Robot.} The real-world setup for our experiments involves two Universal Robots UR5e manipulators equipped with Robotiq 2F-85 grippers, controlled by two Xbox joysticks for collecting demonstrations. A calibrated front RGB-D Realsense camera provides $640\times480$ resolution images at $30$ Hz for observation. We collect $30$ real-world human demonstrations per task for training, while the evaluations are conducted using a Nvidia RTX 4080 GPU. 
% We validate our approach through read-robot experiments on two UR5e robot arms, where a designed base is used to link two arms to form a bimanual manipulation agent. 
% For more setup details, see \Cref{subsec:real_robot_results}.We present two qualitative examples of the action sequences in \Cref{fig:case_study}.

% \textbf{Baselines.} We compare our \method with the state-for-the-art approaches, including PerAct$^2$~\citep{grotz2024peract2benchmarkinglearningrobotic}, which is a strengthened version from the well-known unimanual policy PerAct.
% To exclude the influence of model parameter, we also implement a baseline that incorporates two pre-trained multi-task PerAct~\citep{shridhar2022perceiveractormultitasktransformerrobotic} models. Additionally, we include PerAct-LF in our comparisons, which employs a leader-follower \cite{grotz2024peract2benchmarkinglearningrobotic} architecture using two Perceiver Actor networks. The evaluation metric is the task success rate, which is defined as the percentage of episodes where the agent successfully completes the goal specified by natural language within a budget of $25$ steps.
\noindent\textbf{Baselines.} We compare our \method with the state-of-the-art general bimanual manipulation agents, including the voxel-based method PerAct$^2$~\citep{grotz2024peract2benchmarkinglearningrobotic} and its leader-follower version PerAct-LF, both are modified from the well-known unimanual policy PerAct~\citep{shridhar2022perceiveractormultitasktransformerrobotic}, as well as the multi-view image-based method RVT~\citep{goyal2023rvtroboticviewtransformer}-LF.
To exclude the influence of model parameters, we also implement a counterpart that directly combines two pre-trained PerAct~\citep{shridhar2022perceiveractormultitasktransformerrobotic} policies. 
%%%%%%%%%解释AB不在于两个单臂模型间的通信，所以我们既可以直接接上两个完全独立的单臂模型，也可以接上有通信机制的单臂模型
Note that the proposed method is model-agnostic, which supports different communication mechanisms between single-arm policies, and thus we transfer all $3$ baselines to validate the versatility of \method.
%%%%%%%%%
The evaluation metric is the task success rate, which is defined as the percentage of episodes where the agent successfully completes the instructed goal within $25$ steps.
% specified by natural language

\noindent\textbf{Implementation Details.} 
Following the common training recipe~\citep{shridhar2022perceiveractormultitasktransformerrobotic,grotz2024peract2benchmarkinglearningrobotic,goyal2023rvtroboticviewtransformer}, we use the SE(3) observation augmentation for the expert demonstrations in the training set to improve the robustness of the agents. 
% Specifically, we augment each training sample by perturbing the 3D point cloud with $[\pm0.125m,\pm0.125m,\pm0.125m]$ translation, and rotate it around the $z$-axis by $[0^{\circ},0^{\circ},45^{\circ}]$.
% We also ensure the ground-truth poses of both arms are compatible with the perturbed observation.
For fair comparisons, all compared methods are trained for $100$k iterations on two NVIDIA RTX 3090 GPUs with a total batch size of $4$. 
We use the LAMB optimizer \citep{you2020largebatchoptimizationdeep} with a constant learning rate of $5\times10^{-4}$ to update model parameters, in line with the previous arts~\cite{grotz2024peract2benchmarkinglearningrobotic, shridhar2022perceiveractormultitasktransformerrobotic, goyal2023rvtroboticviewtransformer}.
% For more details about the augmentation strategy and hyper-parameters, please check the supplementary file.
% For more details, please check the supplementary file.
% \Cref{a:se3_data_augmentation} in 

\begin{figure}[t]
    \centering
    \includegraphics[width=0.45\textwidth]{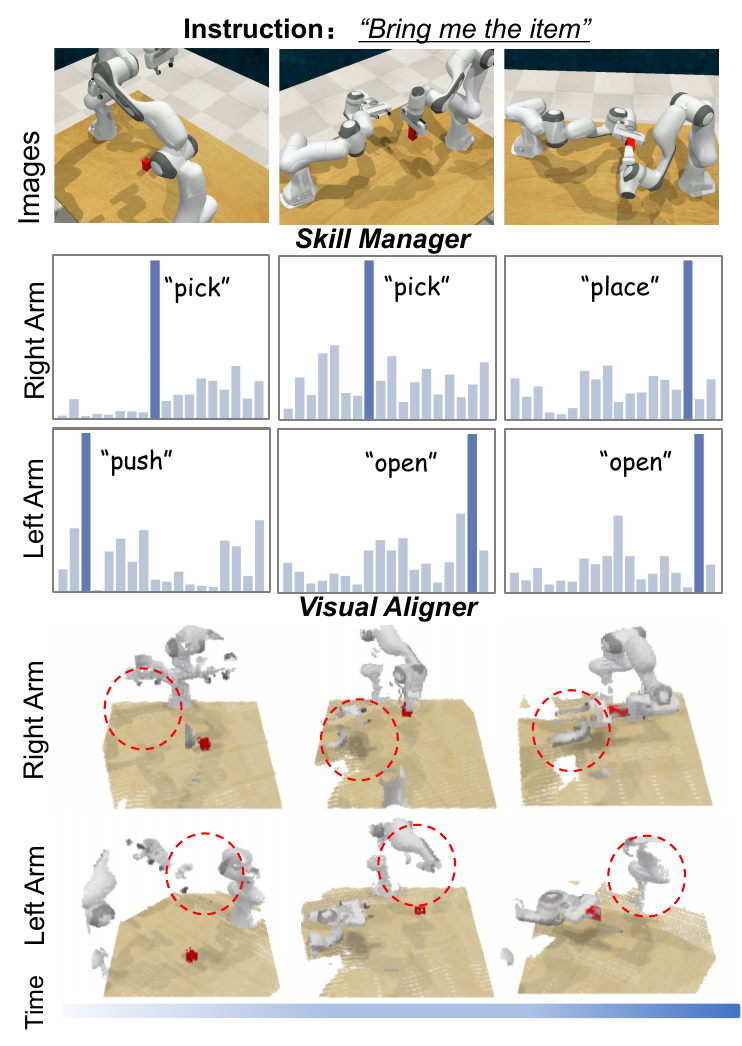}
    \vspace{-0.5cm}
    \caption{\small \textbf{Visualization of \method.} This figure shows in different key timesteps, how the skill manager dynamically schedules skill weights and how the visual aligner decomposes volumetric observation. We use a logarithmic scale for visualization.}
    \label{fig:skill_vis}
    \vspace{-0.5cm}
\end{figure}

\begin{figure}[t]
    \centering
    \includegraphics[width=0.47\textwidth]{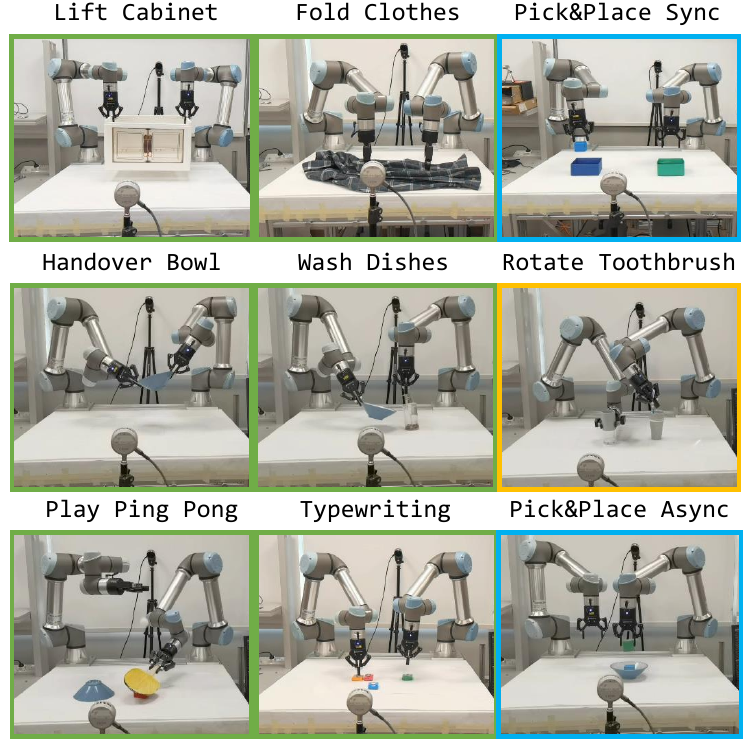}
    % \fbox{
    %     \begin{minipage}[c][0.08\textheight][c]{1.0\textwidth}
    %       \centering{Dummy figure}
    %     \end{minipage}
    %   }
    \caption{\small 
    % \textbf{Real-World Setup and Rollouts.}
    \textbf{Real-World Tasks.}
    The real-world experiments are performed in a tabletop setup with objects randomized in location every episode. \method can simultaneously conduct $9$ complex real-world bimanual manipulation tasks with one model. Different colors mean different success rates.}
    \label{fig:real_world_rollouts}
    \vspace{-0.5cm}
\end{figure}

\subsection{Comparison with the State-of-the-Art Methods}\label{subsec:comp_with_sota}

In this section, we compare our \method with previous stat-of-the-art approaches on RLBench tasksuite. \Cref{table:results} presents a comparison of the average success rates for each task and the average performance is shown in \Cref{fig:teaser}. Our method achieves the highest overall performance, with an average success rate of $32.00$\%, setting a new state-of-the-art in general bimanual manipulation.
% \textbf{Benefits. }
% A cutting-edge bimanual method is PerAct$^2$, which is built upon the well-recognized unimanual PerAct framework, leveraging a shared latent space to enable the cooperation of robot arms.
% Despite its novel bimanual architecture, PerAct$^2$ does not fully exploit the generalization capabilities demonstrated by unimanual models, such as PerAct, which have been highly effective across various manipulation tasks. 
% The lack of knowledge transfer from these effective unimanual policies limits the performance of PerAct$^2$ in complex bimanual tasks. 
% However, the shared latent space of PerAct$2$ struggles to gain enough generalizability due to the scarcity of bimanual demonstrations.
% In contrast, 
\method leverages the knowledge distilled from unimanual models and successfully transfers it to guide general bimanual manipulation. 
% This strategic integration enables more precise and context-aware action prediction. 
% \textbf{Advantage in Long-Horizon Tasks. }
% We further observe that our \method demonstrates a more significant improvement in long-horizon, multi-stage tasks compared to simpler, short-horizon tasks.  For instance, in tasks such as \texttt{handover} and \texttt{oven}, which require continuous coordination between both arms over a longer period, our method shows a more pronounced performance boost compared to 0\% success rate of PerAct$^2$. This can be attributed to \method’s ability to dynamically manage and adapt skill combinations over time, which is particularly crucial in tasks with evolving scene dynamics.  
% In contrast, for simpler tasks like \texttt{push buttons}, the performance improvement is less dramatic, as these tasks rely more on straightforward manipulation. This phenomenon suggests that \method is especially effective in complex, longer-horizon tasks where adaptive action prediction and coordination are critical. 
% It reinforces the idea that transferring knowledge from unimanual models, combined with the skill managing mechanism, allows for more flexible and precise action generation in diverse and challenging environments.\\
As a result, our method outperforms the cutting-edge bimanual method PerAct$^2$ by a significant improvement of $17.33$\% on average.
With a limited set of $20$ demonstrations, our method still defeats the baseline with a sizable margin of $10.50$\%.
Note that we aim to train multi-task agents to handle a wide range of tasks within a single model, hence the performance is lower than the reported in the original paper.
% detailed analysis
Especially, we observe that \method shows greater improvement in long-horizon tasks like \texttt{put in fridge}, multi-variations tasks like \texttt{press buttons}, and tasks demand synchronous coordinations like \texttt{straighten rope}. 
% This phenomenon suggests that \method is especially effective in complex, longer-horizon tasks where adaptive action prediction and coordination are critical. 
% PerAct-LF and RVT
% Moreover, \method enables plug-and-play transferring of an alternative architecture of PerAct-LF and the multi-view image-based policy RVT, which also leads to relative boosts of $25.87$\% ($7.00$\% to $9.00$\%) and $116.65$\% ($4.00$\% to $8.67$\%) on average. 
Moreover, \method enables plug-and-play transferring of PerAct-LF and RVT, which also leads to relative boosts of $72.76$\% ($4.92$\% to $8.50$\%) and $39.41$\% ($10.58$\% to $14.75$\%) on average. 
However, incorporating \method slightly affects the performance on the simple, short-horizon \texttt{Lift ball}, due to the additional complexity in input processing.

\begin{figure*}[t]
% \begin{figure}[t]
    \centering
    \includegraphics[width=1\textwidth]{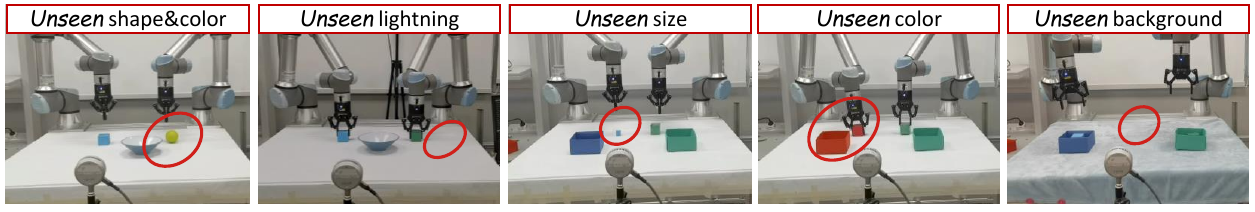}
    \vspace{-0.3cm}
    \caption{\small \textbf{Unseen Data Generalization}. We include multiple distractors in real-world experiments, and find \method generalizes to these settings successfully by unlocking the commonsense held by unimanual base models.}
    \label{fig:real_world_generalization}
    \vspace{-0.3cm}
% \end{figure}
\end{figure*}

% The bimanual agent is composed of two UR5e manipulators.
\subsection{Ablation Study}\label{subsec: ablation_study}

Our \method framework leverages a skill manager to dynamically coordinate unimanual skills, while mitigating the distributional shift between unimanual and bimanual visual inputs with a visual aligner. 
We conduct an ablation study in \Cref{table:ablation} by transferring the powerful unimanul baseline PerAct~\cite{shridhar2022perceiveractormultitasktransformerrobotic} to bimanual tasks. 
% to validate the effectiveness of each component. 
We first implement a vanilla baseline without any of the proposed techniques (Row $1$), where we load the pre-trained PerAct model and directly finetune the baseline to predict bimanual action, which shows enhanced long-horizon task execution and generalizability with an unneglected performance drop in tasks that require proper coordination.

\noindent\textbf{Skill Manager.} 
% By adding a one-hot selection mechanism, the skill manager chooses the skill with the highest predicted probability from the skill set at each timestep. This skill is then represented as a one-hot vector, focusing the agent’s actions on the most relevant skill, the performance improves by 2\% compared to the baseline. We then replace the one-hot selection mechanism with a linear combination of all skills in the skill set, rather than relying on a single skill. 
% By employing the linear combination approach, along with the associated residual, we observe that the average success rate increases by 2.67\% compared to using a single skill, which indicates the enhanced generalization ability of the model across tasks in robotic manipulation. Especially,  in the tasks that require effective skill managing, such as \texttt{Long}, \texttt{Planning}, \texttt{Motion} and \texttt{Occlusion}, it outperforms the vanilla version by sizable margins, which demonstrates the skill manager’s effectiveness in managing complex interactions, handling high variability, and adapting to long-horizon tasks within dynamic environments.\\
By employing the skill manager in the experiment, we observe that the average success rate increases by $8.92$\% (Row $4$ \vs Row $2$). Notably, in tasks requiring long-term manipulation within the \texttt{Long} category, it significantly outperforms the vanilla version, demonstrating the skill manager’s effectiveness in long-horizon tasks.
% handling

\noindent\textbf{Visual Aligner.} Additionally, we incorporate the spatial soft masking from the visual aligner, resulting in a substantial performance improvement of 3.00\% (Row $3$ \vs Row $2$). Although the inclusion of spatial soft masking mechanism may slightly affect performance in simpler tasks due to increased input processing complexity, it leads to significant gains in overall task success. The gains are even more notable in \texttt{Generalized} and \texttt{Sync} tasks, which demand a high degree of adaptability to scene variations and synchronization for cooperative actions. 
% This demonstrates its ability to effectively manage complex task requirements, enabling precise adjustments to dynamic environments and facilitating seamless coordination between the arms.  % ?
The results demonstrate the visual aligner preserves the generalizability stored in the unimanual policy perfectly, facilitating seamless coordination between the arms.
% Conclude
By integrating all techniques in our \method, the success rate improves from $14.67$\% to $32.00$\% (Row $5$), validating the importance of properly mining unimanual knowledge for bimanual manipulation.

% unimanual knowledge and 3D scene soft mask to achieve superior performance in bimanual robotic manipulation.
% \begin{table*}[t]
%   \centering
%   \scriptsize
%     \caption{\small \textbf{Comparison of Our Methods with Different Techniques.} We manually categorize the $12$ RLBench2 task to $6$ groups for further interpretability. For more details regarding the classification method, please refer to the supplementary file.}
%   \setlength{\tabcolsep}{3pt} 
%     \begin{tabular}{c|ccc|cccccc|c}
%     \toprule
%     Row ID & One-hot Skill. & Skill Manager. & Visual Aligner. & \texttt{Long} & \texttt{Planning} & \texttt{Tools} & \texttt{Motion} & \texttt{Lift} & \texttt{Occlusion} & \textbf{Average} \\
%     \midrule
%     1 & \no & \no & \no & 12& 64 & \underline{8} & 12 & 12 & 0 & 14.33\\
%     2 & \yes & \no & \no & \underline{16}& 72 & 4 & 20 & 12 & \textbf{8} & 16.33\\
%     3 & \no & \yes & \no & \textbf{18}& \underline{76} & 4 & \textbf{36} & 15 & \underline{4} & \underline{19.00}\\
%     4 & \yes& \no& \yes& 14 & \textbf{84} & 6 & 20 & \underline{16} & \textbf{8} & 18.67\\
%     \rowcolor{\ourcolor}
%     5 & \no & \yes & \yes & \textbf{18}& \textbf{84} & \textbf{10} & \underline{24} & \textbf{17} & \textbf{8} & \textbf{21.67}\\
%     \bottomrule
%     \end{tabular}
%   \label{table:ablation}
% \end{table*}

\subsection{Qualitative Analysis}\label{subsec:qualitative}

% \noindent\textbf{Visualization of Skill Manager and Visual Aligner.}  
In \Cref{fig:skill_vis}, we visualize linear combination of the skill set and the decomposition of volumetric observation in one complete task to further interpret \method. For illustration, we use $18$ task embeddings from PerAct~\citep{shridhar2022perceiveractormultitasktransformerrobotic} as the initial skill set and explicit soft mask. 
At timestep $1$, the right arm primarily follows the \texttt{pick blocks in sorter} task ($skill\ 7$) from the skill set, as the right arm needs to pick up a block from the table at first. The left arm is guided by the \texttt{push button} task ($skill\ 2$), as its motion only involves a downward push without interacting with any objects. 
At timestep $2$, the right arm holds the red block stationary, and the left arm approaches and grasps it, similar to the motion in \texttt{open drawer} task ($skill\ 16$). 
Additionally, the visual aligner effectively decomposes the voxel space, allowing each arm to focus on relevant information. 
% For instance, in the left arm’s voxel output, some details from the right arm are soft-masked, as shown in the circled area in the figure, preserving the commonsense in the unimanual agent for precise bimanual manipulation.
For instance, at timestep $2$, details of the right arm are soft-masked in the left arm's voxel (as shown in the circled area), which allows the left arm only focus on the red cube without interference from the other arm, enabling the left arm to grasp the red cube more effectively.
% spatial awareness and task coordination for effective bimanual manipulation.
% like the gripper, 

\subsection{Real-Robot Results} \label{subsec:real_robot_results}

% \begin{table}[t]
%   \vspace{-1em} 
%   \caption{\small \textbf{Real-world Results.} Average success rates (\%) of our multi-task model trained and evaluated on $9$ real-world tasks.}
%   \setlength\tabcolsep{4pt} 
%   \centering
%   \scriptsize
%     \begin{tabular}{lcccc} 
%     \toprule
%     Task   & \# of Train & \# of Test & keyframe &\textbf{Average} \\ 
%     \midrule
%     \texttt{Lift}& 30& 5 & 3 & 100 \\
%     \texttt{Handover} & 30& 5 & 3 & 100\\
%     \texttt{Pick in one} & 30& 5 & 4 & 100 \\
%     \texttt{Pick in two} & 30& 5 & 3 & 100 \\
%     \texttt{Press} & 30& 5 & 3 & 100 \\
%     \texttt{Ping Pong} & 30& 5 & 3 & 100 \\
%     \texttt{Fold Clothes} & 30& 5 & 3 &100\\
%     \texttt{Toothbrush} & 30& 5 & 4 &20\\
%     \texttt{Typewriting} & 30& 5 & 6 &100\\
%     \bottomrule
%     \end{tabular}
%   \vspace{-0.5em}  
%   \label{table:real}
% \end{table}

\begin{table}[t]
  % \vspace{-1em} 
  \caption{\small \textbf{Real-world Results.} Average success rates (\%) of our multi-task model trained and evaluated on $9$ real-world tasks.}
  \setlength\tabcolsep{2pt} 
  \centering
  \scriptsize
    \begin{tabular}{lccccc|>{\columncolor{\ourcolor}}c} 
    \toprule
    Task  & Sync. & Object & Variation & Episode & Keyframe &\textbf{Average} \\ 
    \midrule
    \texttt{Lift Cabinet} & \yes & 1 & 1 & 5 & 3.0 & 100.0 \\
    \texttt{Fold Clothes} & \yes & 1 & 1 & 5 & 3.0 &100.0 \\
    \texttt{Pick\&Place Sync} & \yes & 7 & 3 & 15 & 3.0 & 80.0 \\
    \texttt{Handover Bowl} & \no & 1 & 1 & 5 & 4.0 & 100.0 \\
    \texttt{Wash Dishes} & \no & 2 & 1 & 5 & 4.0 & 100.0 \\    % press handsan
    \texttt{Play Ping Pong} & \no & 3 & 1 & 5 & 3.0 & 100.0 \\
    \texttt{Rotate Toothbrush} & \no & 3 & 1 & 5 & 4.0 &20.0\\
    \texttt{Typewriting} & \no & 4 & 1 & 5 & 6.0 & 100.0 \\
    \texttt{Pick\&Place Async} & \no & 5 & 3 & 15 & 4.0 & 80.0 \\    
    \bottomrule
    \end{tabular}
  \vspace{-0.3cm}
  \label{table:real}
\end{table}

% duration, test, items, variations 

% (\texttt{Lift}, \texttt{handover}, \texttt{press}, \texttt{pick in two} and \texttt{pick in one})
We further validate our approach through real-robot experiments. 
% on a bimanual system composed of two UR5e manipulators. 
% Universal Robots 
% For setup details, refer to \Cref{subsec:real-setup}. 
We train one multi-task \method agent that transfers the unimanual policy PerAct~\cite{shridhar2022perceiveractormultitasktransformerrobotic} to $9$ real-world tasks and report the average success rates on \Cref{table:real}.
Guided by ~\cite{taxonomy}, these tasks are designed to cover different challenge levels, such as synchronous and asynchronous coordination, short and long-horizon execution, etc. In the real-robot experiments, our keyframes are manually extracted, providing more stable task guidance compared to heuristic extraction methods in simulation.
The overall success rate across $65$ test episodes in total is $84.62$\%, which illustrates a significant practicality of \method in real-world settings.
Especially, in multi-variant tasks that require high generalizability like \texttt{Pick\&Place Sync} and \texttt{Pick\&Place Async} shown in \Cref{fig:real_world_generalization}, our \method completes $80$\% tasks of different initial positions, object colors, object size, lightning and backgrounds successfully.
The failure cases occur on tasks that demand precise rotation, \eg, \texttt{Rotate Toothbrush}, which could be mitigated by leveraging high-capacity unimanual policy \cite{kim2024openvlaopensourcevisionlanguageactionmodel} or balancing the weight of the rotation term in behavior cloning.
% each task utilizes 30 demonstrations for training.
% Consistent with the simulation results, AnyBimanual achieved over xx\% success on short-horizon tasks like lifting the box. 
% The most frequent failures involved [examples]. 
% This issue could be mitigated by scaling up the expert data to include a broader range of tasks and variations.

% \textbf{Real-World Experiments. }

\section{Conclusion}
\label{sec:conclusion}

In this paper, we have introduced \method, a framework designed to transfer pretrained unimanual manipulation policies to multi-task bimanual manipulation with few bimanual demonstrations. We develop a skill manager to dynamically schedule skill primitives discovered from unimanual policies, enabling their effective adaptation for bimanual tasks. To address the observation discrepancies between unimanual and bimanual systems, we propose a visual aligner that generates spatial soft masks, aligning the visual embeddings of each arm with those used during the pretraining stage of the unimanual policy model. Extensive experiments across $12$ simulated and $9$ real-world tasks demonstrate the effectiveness of \method.
% achieving a $40\%$ relative improvement in average success rate while requiring only $20\%$ of the demonstrations compared to previous state-of-the-art methods. 
% The limitations of our approach primarily arise from 
% the need for careful visual alignment between unimanual and bimanual systems to ensure robust policy transfer across tasks.
% Though \method enables learning robotic bimanual manipulation tasks from few demonstrations, the need of demonstration is still highly-desired to achieve high task completion rate. 
% The limitations of the proposed method is discussed in the supplementary file.
The limitations are discussed in the supplementary file.

% with the addition of embodiment-specific compensations

{
    \small
    \bibliographystyle{ieeenat_fullname}
    \bibliography{main}
}

\newcommand{\blank}{\rule{0.3cm}{0.25mm}~}

\clearpage
\setcounter{page}{1}
\maketitlesupplementary

\section{Additional Experimental Details}
\subsection{Simulation}

\begin{figure}[t]
    \centering
    \includegraphics[width=0.48\textwidth]{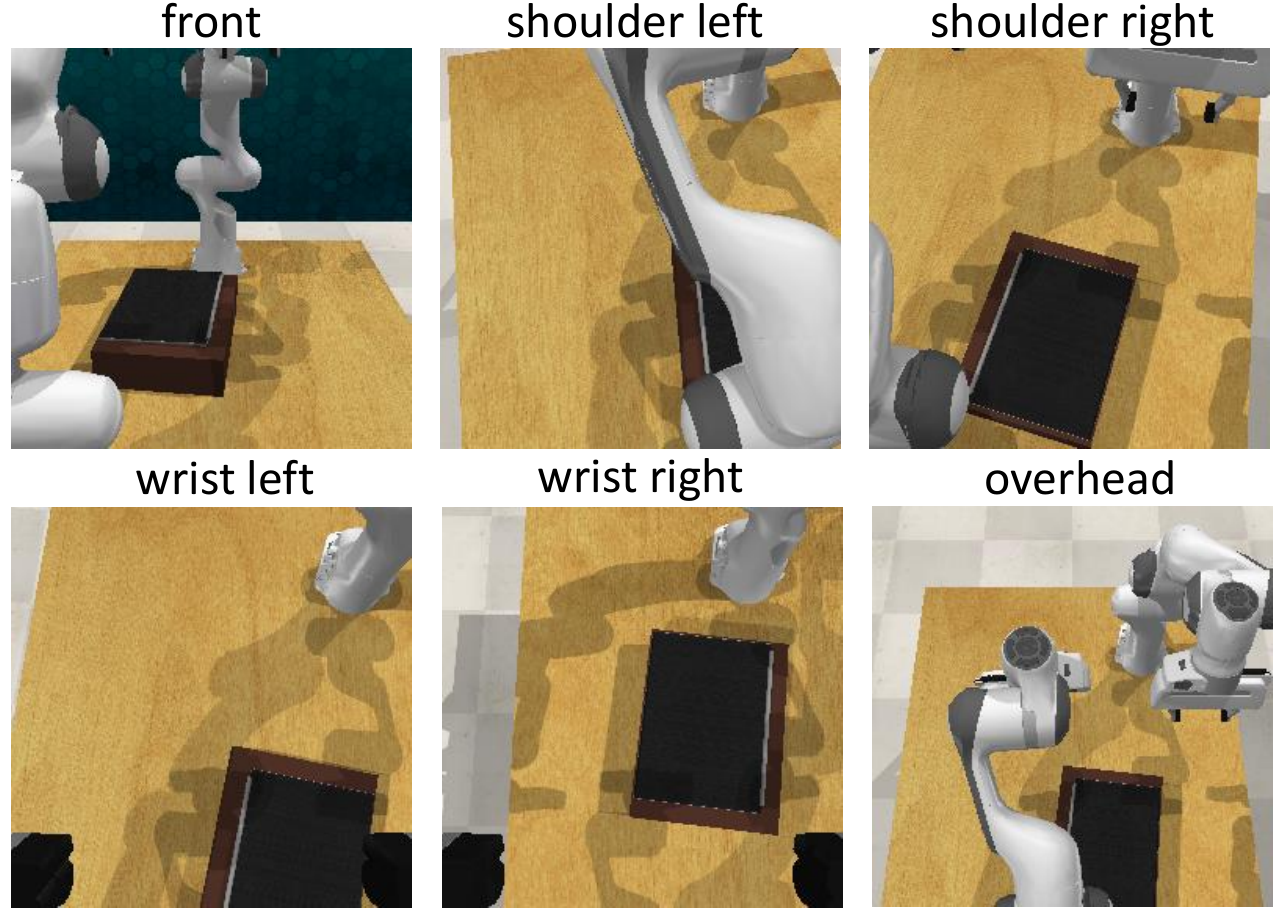}%
    \caption{\small \textbf{Visual Observation in RLBench2.} We adopt $6$ RGB-D cameras to cover the whole workspace.}
    \label{fig:rlbench2_observation}
\end{figure}

We utilize RLBench2 \cite{grotz2024peract2benchmarkinglearningrobotic} as our main simulated task suite for mulit-task learning.
Table~\ref{table:task desc} is an overview of the $12$ selected tasks we use in the experiments. 
Table~\ref{table:single task desc} is an overview of the $18$ selected tasks from RLBench \cite{james2019rlbenchrobotlearningbenchmark} used to pretrain the unimanual checkpoint. 
The task variations include randomly sampled colors, sizes, counts, placements, and categories of objects. 
% We have a color palette of $20$ shades, including red, maroon, lime, green, blue, navy, yellow, cyan, magenta, silver, gray, orange, olive, purple, teal, azure, violet, rose, black, and white. 
% The size of the objects is categorized into two types: short and tall. The number of objects can be either \num{1}, \num{2}, or \num{3}. 
Other properties vary depending on the specific task.
% For instance, there are \num{3} possible placement locations in \texttt{open drawer} task: top, middle and bottom. 
Furthermore, objects are randomly arranged on the tabletop within a certain range, adding to the diversity of the tasks.
The visual observation includes $6$ RGB-D frames of $256 \times 256$ resolutions (\ie, front, shoulder left, shoulder right, wrist left, wrist right and overhead), which are shown in \Cref{fig:rlbench2_observation}.
% adopt the task classification from~\cite{guhur2023instruction} to
In the ablation study, we group the RLBench2 tasks of~\Cref{table:task desc} into $3$ categories according to their key challenges. The task groups include:
\begin{itemize}
    \item The \texttt{Long} group includes long-term tasks that requires more than \num{6.5} keyframes. The included tasks are: \texttt{pick plate}, \texttt{pick laptop}, \texttt{put in fridge}, \texttt{handover item}, \texttt{sweep to dustpan}, \texttt{take out of tray} and \texttt{handover easy}.
    \item The \texttt{Generalized} group includes tasks with multiple variations, which are differentiated by instructions. The included tasks are: \texttt{press buttons} and \texttt{handover item}.
    \item The \texttt{Sync} group requires precise coordination, which often causes failures due to asynchronous manipulation. The included tasks are: \texttt{lift ball}, \texttt{lift tray}, \texttt{straighten rope} and \texttt{push box}.
\end{itemize}

\begin{table*}[t]
\caption{\small \textbf{Unimanual Tasks.} This table shows the $18$ unimanual tasks in RLBench that are used to pretrain the unimanual policy.}
\label{table:single task desc}
\centering
\scriptsize
\setlength\tabcolsep{15pt} 
\begin{tabular}{lccl} 
\toprule
Task                      & \# of Variations  & \# of Average Keyframes       & Human Instruction     Template         \\
\midrule
\texttt{open drawer}      &           3          &           3.0        & \textit{``open the \blank drawer''} \\
\texttt{slide block}      &           4          &           4.7        & \textit{``slide the block to \blank target''} \\
\texttt{sweep to dustpan} &           2          &           4.6        & \textit{``sweep dirt to the \blank dustpan''} \\
\texttt{meat off grill}   &           2          &           5.0        & \textit{``take the \blank off the grill''} \\
\texttt{turn tap}         &           2          &           2.0        & \textit{``turn \blank tap''} \\
\texttt{put in drawer}    &           3          &          12.0        & \textit{``put the item in the \blank drawer''} \\
\texttt{close jar}        &          20          &           6.0        & \textit{``close the \blank jar''} \\
\texttt{drag stick}       &          20          &           6.0        & \textit{``use the stick to drag the cube onto the \blank target''} \\
\texttt{stack blocks}     &          60          &          14.6        & \textit{``stack \blank \blank blocks''}  \\
\texttt{screw bulb}       &          20          &           7.0        & \textit{``screw in the \blank light bulb''} \\
\texttt{put in safe}      &           3          &           5.0        & \textit{``put the money away in the safe on the \blank shelf''} \\
\texttt{place wine}       &           3          &           5.0        & \textit{``stack the wine bottle to the \blank of the rack''} \\
\texttt{put in cupboard}  &           9          &           5.0        & \textit{``put the \blank in the cupboard''} \\
\texttt{sort shape}       &           5          &           5.0        & \textit{``put the \blank in the shape sorter''} \\
\texttt{push buttons}     &          50          &           3.8        & \textit{``push the \blank button, [then the \blank button]''} \\
\texttt{insert peg}       &          20          &           5.0        & \textit{``put the ring on the \blank spoke''} \\
\texttt{stack cups}       &          20          &          10.0        & \textit{``stack the other cups on top of the \blank cup''} \\
\texttt{place cups}       &           3          &          11.5        & \textit{``place \blank cups on the cup holder''} \\
\bottomrule
\end{tabular}
\end{table*}

\begin{table*}[t]
\caption{\small \textbf{Bimanual Tasks.} This table shows the $12$ bimanual tasks in RLBench2 that are used to fine-tune the bimanual policy.}
\label{table:task desc}
\centering
\scriptsize
\setlength\tabcolsep{15pt} 
\begin{tabular}{lccl} 
\toprule
Task                       & \# of Variations  & \# of Average Keyframes    & Human Instruction Template  \\
\midrule
\texttt{pick laptop}       & 1              & 7.2            & \textit{``pick up the notebook.''}            \\
\texttt{pick plate}        & 1              & 6.6            & \textit{``pick up the plate.''}               \\
\texttt{straighten rope}        & 1              & 5.9            & \textit{``straighten the rope.''}             \\
\texttt{lift ball}            & 1              & 4.0            & \textit{``lift the ball.''}                   \\
\texttt{lift tray}              & 1              & 5.1            & \textit{``lift the tray.''}                   \\
\texttt{push box}               & 1              & 2.1            & \textit{``push the box to the red area.''}    \\
% \texttt{put item in drawer}     & 3              & 8.4            & \textit{``put the item into the \blank drawer.''} \\
\texttt{put in fridge}   & 1              & 7.8            & \textit{``put the bottle into the fridge.''}  \\
\texttt{press buttons}       & 5              & 4.0            & \textit{``push the \blank and \blank button.''} \\
\texttt{handover item}       & 5              & 7.6            & \textit{``hand over the \blank item.''}      \\
\texttt{sweep to dustpan}         & 1              & 7.3            & \textit{``sweep the dust to the pan.''}       \\
\texttt{take out tray}              & 1              & 8.7            & \textit{``take tray out of oven.''}                   \\
\texttt{handover easy}   & 1              & 7.5            & \textit{``hand over the item.''}              \\
% \texttt{take tray out of oven}  & 1              & 8.7            & \textit{``take tray out of oven.''}           \\
\bottomrule
\end{tabular}
\end{table*}

\subsection{Real-Robot}\label{subsec:real-setup}
\begin{figure}[t]
    \centering
    \includegraphics[width=0.48\textwidth]{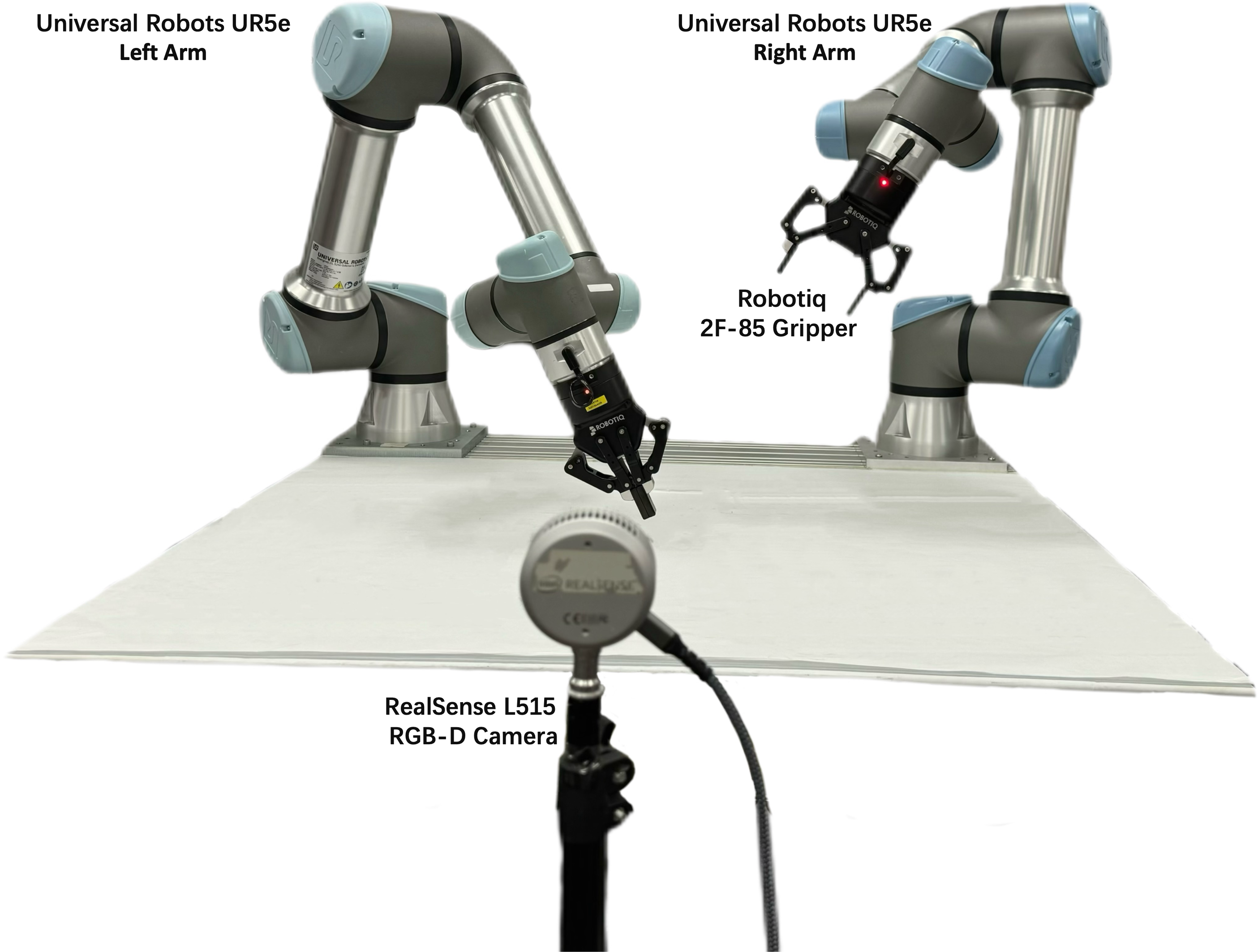}%
    \caption{\small \textbf{Real-Robot Setup} with one RealSense L515 RGB-D Camera and Two UR5e Manipulators. }
    \label{fig:real-rob}
\end{figure}

\textbf{Hardware Setup.} The real robot setup uses two Universal Robots UR5e manipulators, each equipped with a Robotiq 2F-85 gripper. See \Cref{fig:real-rob} for reference. For the perception, we use a Realsense RGB-D camera mounted on a tripod, positioned to mimic the viewpoint of human eyes during bimanual tasks. The Realsense provides RGB-D images at a resolution of 640x480 with a frame rate of 30 Hz. The extrinsics between the camera and right arm base-frame are calibrated using the easy\_handeye package\footnote{\url{https://github.com/IFL-CAMP/easy_handeye}}. Additionally, an ARUCO marker\footnote{\url{https://github.com/pal-robotics/aruco_ros}} attached to the UR5e’s end-effector is employed to aid in the calibration process.

\begin{figure*}[t]
    \makebox[\textwidth][c]{%
        \includegraphics[width=1.0\textwidth]{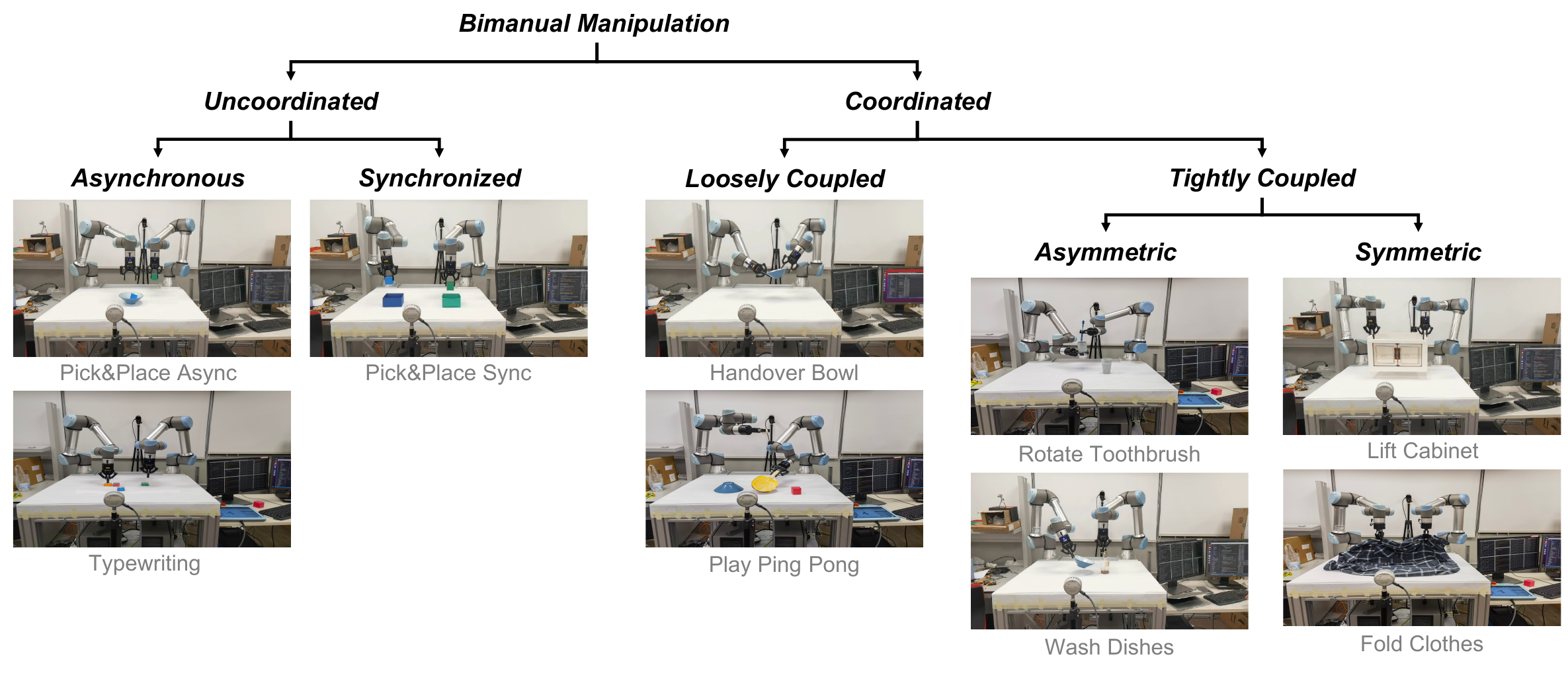}%
    }
    \caption{\small \textbf{Real-Robot Task Taxonomy}. Following~\citep{taxonomy}, we classify the reported $9$ real-world tasks into $5$ categories according to their collaboration patterns.}
    \label{fig:taxonomy}
    \vspace{-0.2cm}
\end{figure*}

\noindent\textbf{Tasksuite Descriptions.} 
We provide a detailed description of the real-world tasks we used to evaluate our \method in \Cref{fig:taxonomy}, where we follow~\citep{taxonomy} to classify our $9$ real-world experiments into $5$ categories, covering $5$ bimanual collaboration patterns. 
\begin{itemize}
    \item \texttt{Lift Cabinet} requires the agent to simultaneously hold two sides of the cabinet to lift it up, which poses challenges to synchronous coordination.
    \item \texttt{Fold Clothes} requires the agent to simultaneously grasp and fold the flexible clothes, which involves soft manipulation and synchronous coordination.
    \item \texttt{pick\&place Sync} requires the agent to simultaneously pick up two colored cubes specified by the human instructions, and then place them in corresponding boxes, which involves semantic understanding.
    \item \texttt{Handover Bowl} requires the agent to lift a bowl and hand it over to the other hand, posing challenges in coordinating the handover between hands.
    \item \texttt{Wash Dishes} requires the agent to pick up a bowl and then apply detergent to it, testing the asynchronous coordination of both hands.
    \item \texttt{Play Ping Pong} requires the agent to pick up both a ball and a racket, toss the ball with one hand and hit it with the racket in the other in a high toss serve, testing asynchronous coordination.
    \item \texttt{Rotate Toothbrush} requires the agent to pick up a toothbrush and a cup simultaneously, then rotate the toothbrush 180 degrees to drop it into the cup, challenging the precision of rotation and coordination of both hands.
    \item \texttt{Type Writing} requires the agent to sequentially type the letters "robot" on a keyboard using both hands, testing long-horizon coordination.
    \item \texttt{Pick \& Place Async} requires the agent to simultaneously pick up two cubes of different colors and then place them into the same bowl, testing the asynchronous coordination of both hands.
\end{itemize}

\begin{figure}[t]
    \centering
    \includegraphics[width=0.48\textwidth]{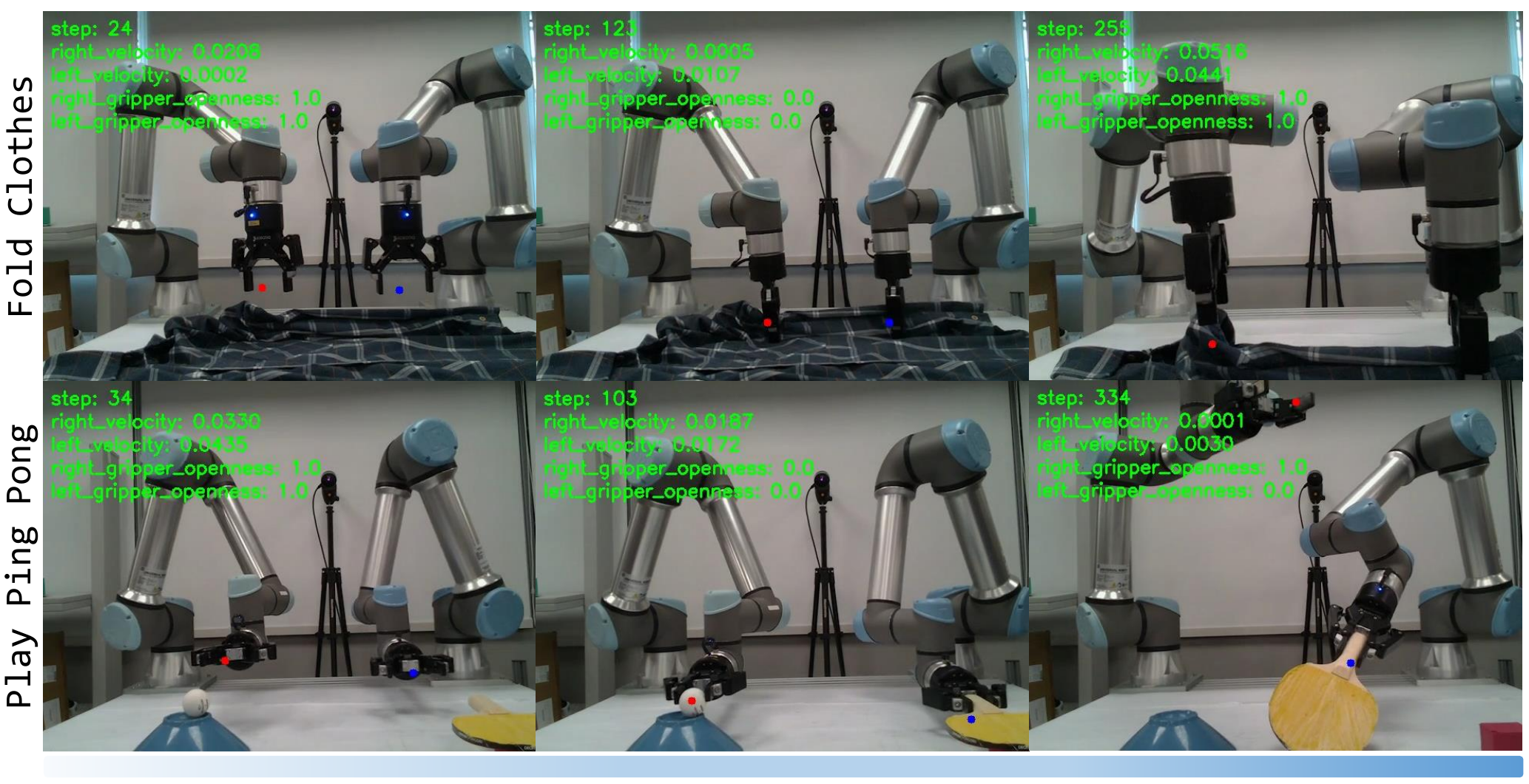}%
    \caption{\small \textbf{Real-world Keyframes.} We manually select keyframes from the collected trajectories to simplify training.}
    \label{fig:keyframe}
    \vspace{-0.2cm}
\end{figure}

\noindent\textbf{Data Collection and Preprocessing.} We collect demonstrations with two Xbox joysticks. Each gamepad is a 6-DoF controller. The gamepads adjust the velocity of the arm’s end-effector to translate and rotate in all directions, with reference to the arm's base frame. 
% For motion planning, we utilized the Universal Robots ROS Driver. 
For robot control, we utilize the Universal Robots ROS Driver\footnote{\url{https://github.com/UniversalRobots/Universal_Robots_ROS_Driver}}. 
After collecting the complete trajectory, we extract keyframes from the trajectory manually to simplify the robot learning. Two examples of the extracted keyframes are shown in \Cref{fig:keyframe}.
The total time consumption of collecting one episode including controlling the robot arms and keyframe extraction is $\sim1.5$ minutes.

\begin{figure}[t]
    \centering
    \includegraphics[width=0.48\textwidth]{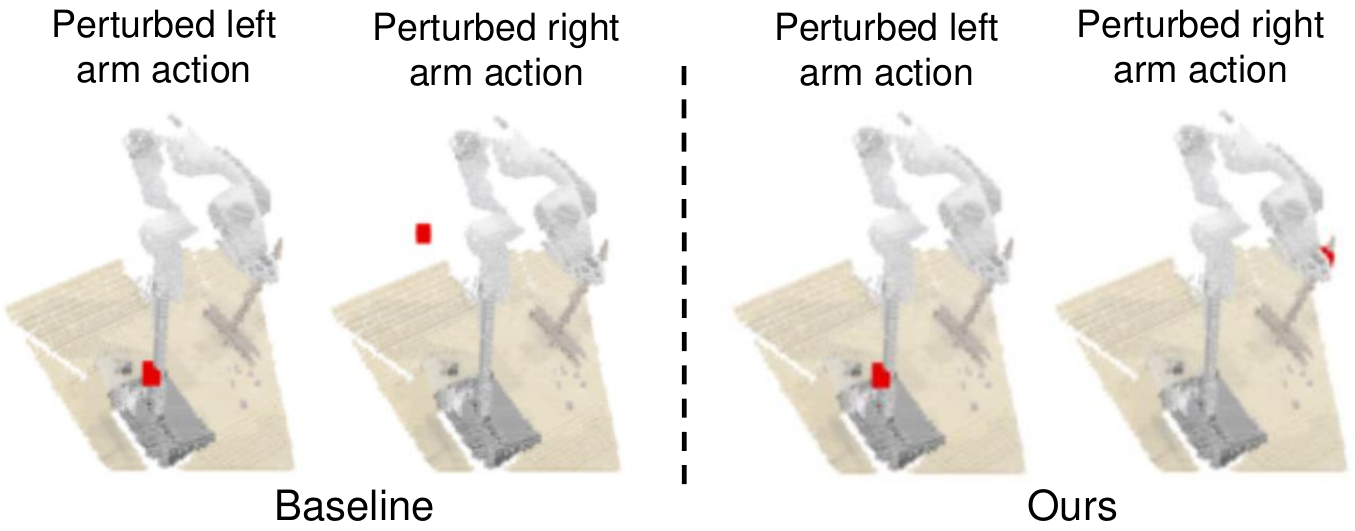}
    \caption{\small \textbf{SE(3) Augmentation.} We slightly modify the SE(3) augmentation strategy to ensure that the perturbed ground-truth actions of both arms are reasonable.
    }
    \label{fig:se3}
\end{figure}

\noindent\textbf{Data Augmentation.}\label{a:se3_data_augmentation} In line with~\citep{shridhar2022perceiveractormultitasktransformerrobotic, grotz2024peract2benchmarkinglearningrobotic}, we turn on SE(3) augmentation to boost the robustness of all learned policies. Based on the open source code of~\citep{grotz2024peract2benchmarkinglearningrobotic}, we modify the perturbation strategy to ensure that the perturbed ground-truth actions of both arms are reasonable in \Cref{fig:se3}. We plan to release our code for more details.
% However, we find bugs in

% \noindent\textbf{Training and Execution.} We train an AnyBimanual agent using 100 demonstrations for each task, incorporating translational and rotational perturbations into the training samples to enhance the model’s robustness. Each task undergoes training for a full day on a single Nvidia RTX 3090 GPU with a batch size of 2. During the evaluation phase, we select the final checkpoint form training, as there is no effective method available for assessing the model’s performance during the training period. We perform inference on a single Nvidia RTX 4080 GPU.\\

\noindent\textbf{Policy Deployment.}
During the testing phase, we first randomize the positions of related objects in the current task. After resetting the manipulators to the home positions, we query the policy with the language instruction, initial observation and proprioception to obtain the next best poses
of two end-effectors.
We then use Moveit\footnote{\url{https://github.com/moveit/moveit}} to reach the poses and query the policy again with new observations, until task completion or reaching the maximal episode length.

\section{Additional Implementation Details}

% TODO
\subsection{Hyperparameters}\label{subsec:hyper-paramters}
The hyperparameters used in \method are shown in Table~\ref{table:hyperparam}. 
% To train the robotic manipulation agent, we use $\lambda_{  \text{Geo}}=0.01$, $\lambda_{  \text{Sem}}=0.0001$ and $\lambda_{  \text{Dyna}}=0.001$ to focus on the action prediction. 
To ensure that the auxiliary objectives $\mathcal{L}_{\text {skill}}$ and $\mathcal{L}_{\text {voxel}}$ remain in the same magnitude as $\mathcal{L}_{\text {BC}}$, we set $\lambda_{\text{skill}} = 0.0001 $ and $\lambda_{\text{voxel}} = 0.001$.
Other hyperparameters are in line with previous works~\citep{shridhar2022perceiveractormultitasktransformerrobotic, grotz2024peract2benchmarkinglearningrobotic} for fair comparisons. 

% RVT, PerAct2共享？
\begin{table}[t]
\caption{\small\textbf{A default set of hyper-parameters.}}
\centering
\label{table:hyperparam}
\scriptsize
\setlength\tabcolsep{10pt} 
\begin{tabular}{l  l }
\toprule
Config  & Value \\
% \shline
\midrule
Training iteration & $100$k \\
Image size & $256\times 256 \times 3$\\
Voxel size & $100\times 100 \times 100$\\
Batch size & $4$ \\
Optimizer & LAMB~\citep{you2020largebatchoptimizationdeep}\\
Learning rate & $0.0005$ \\
Weight decay & $0.000001$ \\
% Number of transformer blocks & $6$ \\
% Number of latents in Perceiver Transformer & $2048$\\
% Dimension of CLIP language features & $512$\\
$\lambda_{\text{skill}}$ & 0.0001 \\
$\lambda_{\text{voxel}}$ & 0.001 \\
Number of skill primitives $K$ & 18 \\
Dimension of Skill Representations $D$ & 512 \\
% Max Length of Tokens & 77 \\
\bottomrule
\end{tabular}
\end{table}

\begin{figure*}[t]
    \centering
    \includegraphics[width=0.95\textwidth]{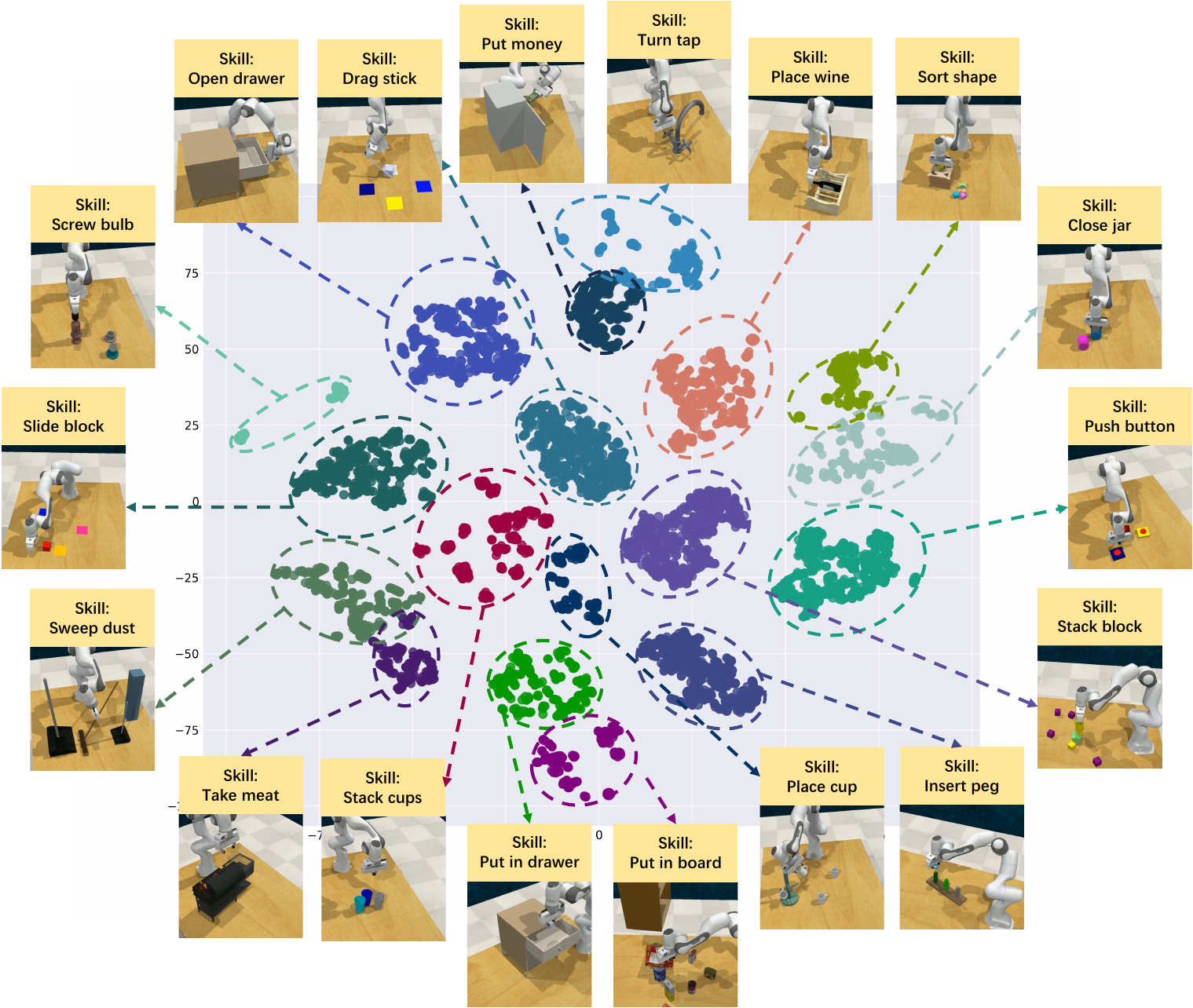}
    \caption{\small \textbf{Skill Clustering.} We cluster the skill representations to further interpret the discovered skills of our \method.
    }
    \label{fig:skill_clustering}
    \vspace{-0.2cm}
\end{figure*}

% 感觉可以不放了？
\begin{figure*}[t]
    \centering
    \includegraphics[width=1\textwidth]{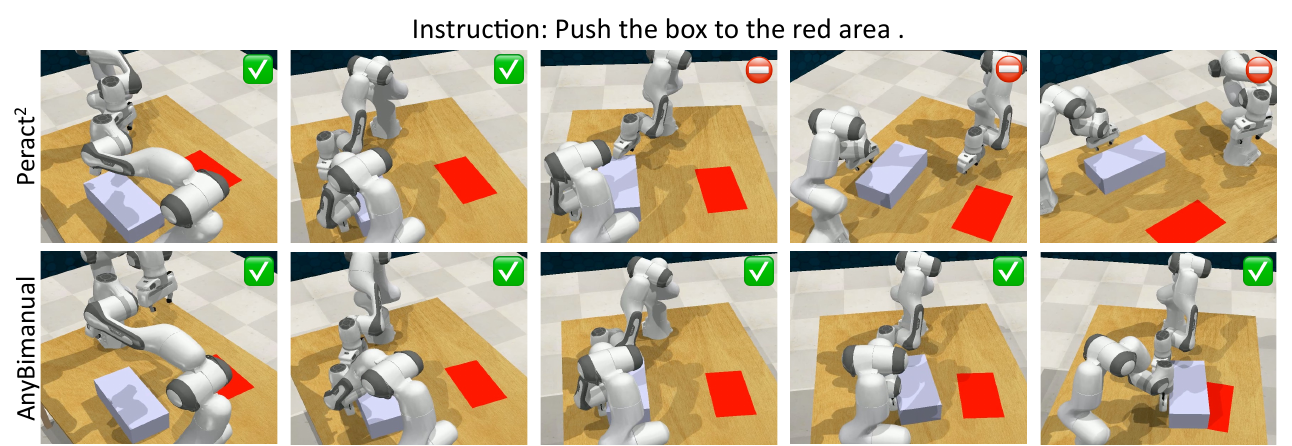}
    \caption{\small \textbf{Comparions of the rollouts of PerAct$^2$ and our \method.} 
    % Compared to the state-of-the-art method PerAct$^2$, \method demonstrates superiority in tasks that demand complex collaboration patterns by incorporating the skill manager and the visual aligner to correctly schedule the manipulators.
    \method demonstrates complex collaboration patterns by incorporating the skill manager and the visual aligner to schedule the manipulators correctly.
    }
    % \vspace{-0.4cm}
    \label{fig:case_study}
\end{figure*}

\begin{figure*}[t]
    \centering
    \includegraphics[width=1\textwidth]{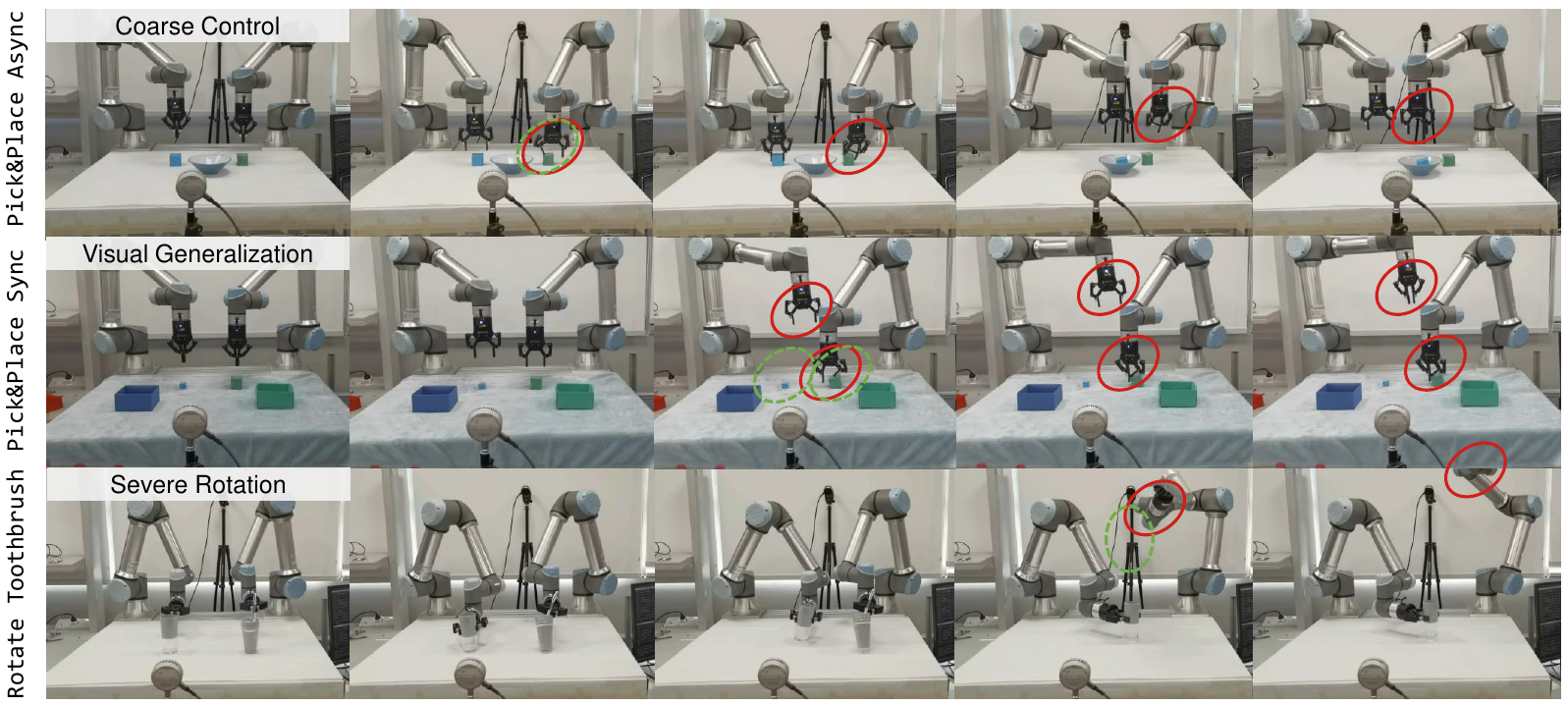}
    \caption{\small \textbf{Error Modes in Real-world Rollouts.} We study the common failure cases of our \method in real-world settings, where the incorrect end-effector actions are marked with red circles. We also provide the correct actions with green circles for reference.
    }
    \label{fig:error_modes}
    \vspace{-0.2cm}
\end{figure*}

\subsection{Architectures}

\noindent\textbf{Skill Manager.}
We fuse voxel embeddings and language embeddings by projecting both to a hidden size of $256$ and concatenating them. After layer normalization, the combined sequence is processed by a frozen GPT-2 Transformer configured with a hidden size of $256$ and $8$ attention heads. The transformer's output is passed through a linear layer to produce logits over 18 classes, and softmaxed to obtain probabilities. These probabilities are then used to reconstruct the skill representation via a weighted sum over the skill primitives.

\noindent\textbf{Visual Aligner.}
We employ a lightweight module to process input embeddings through a Conv1D-ReLU block, then split into two branches to generate left and right masks via additional Conv1D-ReLU layers. After element-wise multiplication of these masks with the original input, we incorporate residual connections by adding the masked outputs back to the input embeddings.

\subsection{Skill Primitives}
The skill primitives use CLIP embeddings of unimanual language templates (\eg, `open the \blank drawer') from Table 1 of the supplementary file, as we expect them to represent foundational motions that generalize across object categories in unseen scenarios.
% \section{Additional Quantitative Analysis}
% \label{a:additional_quantitative}

\section{Additional Experimental Results}

% \subsection{Additional Ablation Studies}\label{subsec:additional_ablation}
% In this subsection, we investigate the influences of the choice of different techniques in our \method.

\subsection{Skill Clustering}\label{subsec:skill_clustering}
The goal of the skill manager is to obtain the language embedding via linear skill representations, so that the pre-trained unimanual policy model (\ie, PerAct) can be prompted to generate feasible manipulation actions. 
In \Cref{fig:skill_clustering}, we cluster the skill representations while evaluating the last checkpoint of PerAct+\method in RLBench2. 
We can observe that the learned skill representations are reasonably clustered into $18$ categories, which corresponds to the number of pretrained unimanual tasks. 
We also visualize the related unimanual skill of each cluster by finding the nearest unimanual task embedding.
The clustering results further interpret that the proposed skill manager enables compact skill representation learning with rich semantics for dynamic scheduling.

% \noindent\textbf{Comparisons of Rollouts.} We present two qualitative examples of the generated action sequences in \Cref{fig:case_study}, comparing PerAct$^2$ and our \method method. 
% In the top case, the agent is instructed to "Bring me the item". The results show that the PerAct$^2$ agent's right arm incorrectly identifies the position of the red block, causing the gripper to miss the block after its initial downward motion. When the right arm attempts to grab the block again, it pushes the block over, leading to a failed action sequence. In contrast, our \method agent accurately identified the position of the red block and successfully grabs it, handing it over to the left arm, owing to that our method can correctly understand the scene dynamics of objects in contact.
% new
% In the top case, the agent is instructed to “Lift the box.” The right arm first grasps the box, followed by the left arm securing its hold. Both arms then lift the box simultaneously, demonstrating precise and coordinated bimanual manipulation.
% In the bottom case, the instruction is "Push the box to the red area". The PerAct$^2$ agent fails to touch the box with both arms and only mimics the expert's forward-pushing motion. In contrast, the agent with \method coordinates both arms to contact with the box and successfully pushes it into the red area, demonstrating superior scheduling ability.
% ability in having each arm correctly identify contact with the object and successfully complete the task.
% resulting in an incomplete action

\subsection{Rollout Comparisons}
We present a qualitative example of the generated action sequences in \Cref{fig:case_study}, comparing PerAct$^2$ and our \method method. 
In this case, the instruction is ``Push the box to the red area''. The PerAct$^2$ agent fails to touch the box and only mimics the expert's forward-pushing motion, resulting in an incomplete action. In contrast, our \method agent ensures both arms make contact with the box and successfully push it into the red area, demonstrating the superior ability of our method in having each arm correctly identify contact with the object and successfully complete the task.

\subsection{Visualization}
\begin{figure}[t]
    \centering
    \includegraphics[width=0.44\textwidth]{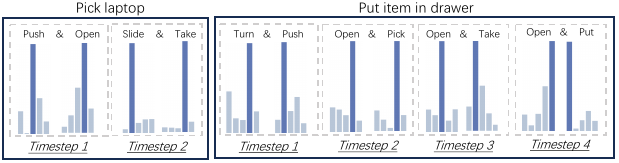}
    \caption{\small \textbf{Visualization of Skill Manager.}}
    \label{fig:111}
    \vspace{-0.2cm}
\end{figure}

\textbf{Skill Manager.} We visualize the skill manager in two challenging tasks requiring heavy coordination. (\Cref{fig:111})
\method completes these tasks by scheduling skill representations across arms simultaneously and dynamically.

% \begin{figure}[t]   % 0.09
%     \centering
%     \includegraphics[width=0.44\textwidth]{pic/skill_manager.pdf}
%     % \fbox{\rule{0pt}{0.4in} \rule{0.45\textwidth}{0pt}}
%     \label{fig:vis_skill}
%     \caption{\small \textbf{Visualization of Skill Manager.}
%     }
%     \vspace{-0.5cm}
% \end{figure}

\begin{figure}[h]   % 0.09
    \centering
    % \fbox{\rule{0pt}{0.4in} \rule{0.45\textwidth}
    % {0pt}}
    \includegraphics[width=0.47\textwidth]
    {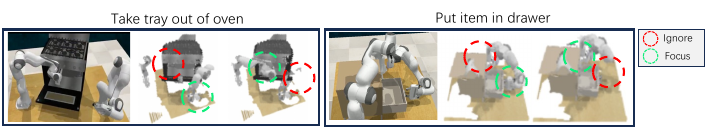}
    % {pic/visual_aligner_v3.pdf}
    \caption{\small \textbf{Visualization of Visual Aligner.}}
    \label{fig:vis_vis}
\end{figure}

\noindent\textbf{Viusal Aligner.} By predicting spatial soft masks for voxel observation, the visual aligner encourages both arms to focus on the interaction and ignore irrelevant parts. (\Cref{fig:vis_vis})

% \subsection{Ablation Study}\label{subsec:ablation_study}

\subsection{Error Modes}\label{subsec:error_mode}
To further study the limitations and risks of the proposed method, we visualize the common failure cases in \Cref{fig:error_modes}. 
Common error modes include coarse control, which is often viewed in \texttt{Pick\&Place Sync} and \texttt{Pick\&Place Async}. 
In the upper case, the robot fails to grasp the green block due to a small end-effector deviation, which may be caused by inconsistent data collection. 
In the middle case, the robot struggles to generalize to different backgrounds and object sizes, which leads to an out-of-distribution movement of the left robot arm.
In the bottom case, the low-level motion planner cannot handle the severe rotation predicted by the agent, which can result in a dangerous cup fall.
We hope that the analysis can enlighten future works in bimanual policy transferring and prevent potential risks in real-world deployments.

% \section{Additional Qualitative Analysis}
% \label{a:additional_qualitative}

\subsection{Bimanual Decomposability}\label{subsec: bi_de}
\begin{figure}[h]
    \centering
    \includegraphics[width=0.4\textwidth]{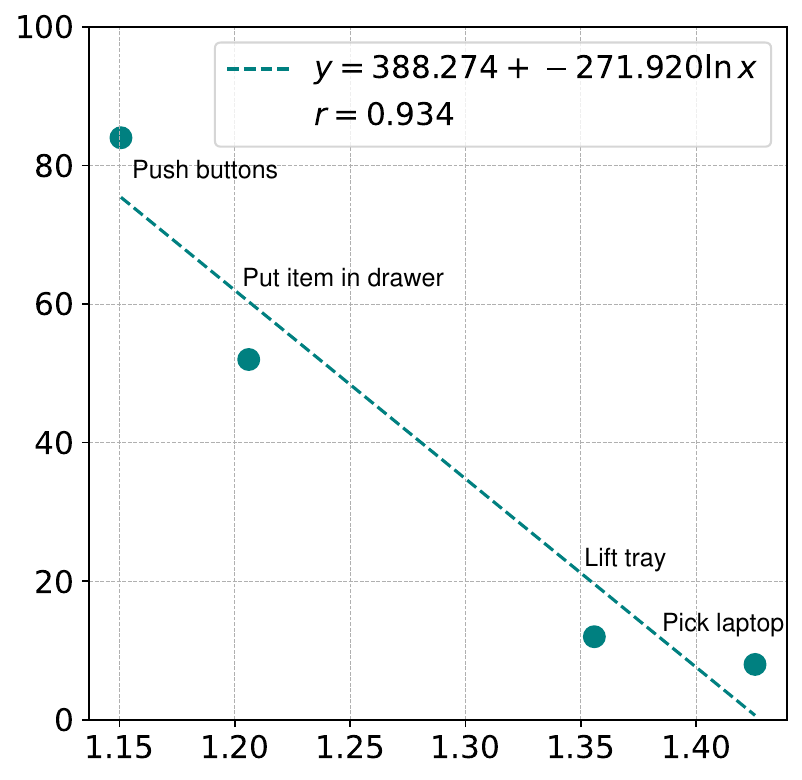}
    \caption{\small \textbf{Bimanual Decomposability.}}
    \vspace{-0.2cm}
    \label{fig:curve}
\end{figure}

We complement a heuristic measurement to assess the feasibility of decomposing a given bimanual task with a learned skill manager at the semantic level.
Specifically, higher entropy in the predicted combination weights indicates greater difficulty in decomposition with certain unimanual primitives. In \Cref{fig:curve}, this metric strongly correlates with final success rates in four simulation tasks, delineating the boundary.

\subsection{Real-world Videos}\label{subsec:video}
We provide $9$ additional comprehensive real-world episodes generated by our \method in the attached video file (\emph{demo.mp4}). Note that all $9$ tasks are completed by a single model with different natural language instructions to specify the current task.

% \noindent\textbf{Real-world Experiments with Few Demonstrations.}
% \subsection{Real-world Experiments with Few Demonstrations.}
\subsection{Real-world Experiments with Limited Data.}
To further validate the effectiveness and practicality of \method with limited expert data, we train \method on $9$ real-world tasks, each with $5$ demonstrations. As per the main paper, we evaluate each variation over $5$ episodes, and report the success rates of each task.
The results are illustrated in \Cref{table:real_few}, where \method still maintains an acceptable average success rate of $53.33$\%.

\begin{table}[t]
  \caption{\small \textbf{Real-world Results with $5$ Demonstrations Per Task.} We train \method on $9$ real-world tasks with $5$ demonstrations for each task, and evaluate each variation over $5$ episodes.}
  \setlength\tabcolsep{20pt} 
  \centering
  \scriptsize
    \begin{tabular}{l>{\columncolor{\ourcolor}}c} 
    \toprule
    Task & \textbf{Success Rate (\%)} \\ 
    \midrule
    \texttt{Lift Cabinet} & 80.0 \\
    \texttt{Fold Clothes} & 80.0 \\
    \texttt{Pick\&Place Sync} & 80.0 \\
    \texttt{Handover Bowl} & 20.0 \\
    \texttt{Wash Dishes} & 20.0 \\
    \texttt{Play Ping Pong} & 40.0 \\
    \texttt{Rotate Toothbrush} & 20.0 \\
    \texttt{Typewriting} & 60.0 \\
    \texttt{Pick\&Place Async} & 80.0 \\    
    \bottomrule
    \end{tabular}
  \label{table:real_few}
\end{table}
% 50k

\section{Discussions on Limitations}
\label{a:limitations}

% \noindent\textbf{Highly Dynamic Tasks.} 
% The proposed method assumes a consistent skill between keyframes, which can last for several seconds. This design may fail to adapt to highly dynamic tasks with unexpected changes. This could be solved by introducing a high-frequency low-level planner \cite{xian2023chaineddiffuser}.

% \noindent\textbf{Keyframe Extraction, Motion Planner, and Latency Matching.} 
% \noindent\textbf{Keyframe Extraction.} 
% Our skill-transferring framework assumes an implicit hypothesis that any task demonstration can be reasonably separated into subtasks and corresponding keyframes. For example, we extract the keyframe by heuristics (one gripper state has changed, or one robot velocity is near zero) in simulation, and by manual determination in the real world (press a `q' key when replaying the demonstration video). 
% However, both the heuristic and manual extraction need additional crafting, inconsistent or redundant keyframes may confuse the policy, and some tasks even can not be simply divided by keyframes. 
% Automatic keyframe extraction methods \cite{shi2023waypointbased, zhou2024maxmi} may help convenient and consistent keyframe identification.

% \noindent\textbf{Dexterous Manipulation.}

\noindent\textbf{Cross-embodiment Skill Transferring.}
\method needs careful visual alignment between unimanual and bimanual systems to ensure robust policy transfer, which means it is not designed for cross-embodiment skill transferring. Even though the 7-DoF end-effector action space can be used with various robots, differences in the visual appearance of unimanual and bimanual robots may limit generalizability. 
Because the visual aligner only attempts to align the observation by segmenting out unimanual workspace from the bimanual system, without considering their appearance differences.
This can be addressed by visual inpainting \cite{wang2023imagen} or transferring cross-embodiment unimanual policies \cite{wang2024scaling}.

\noindent\textbf{Cross-task Generalization.}
Although \method enables general multi-task bimanual manipulation with few demonstrations, it struggles to perform zero-shot generalization to unseen tasks. This is mainly due to the end-to-end skill manager and visual aligner still necessitating expert demonstrations to learn proper coordination patterns and unlock the capacity of unimanual policy. Explicit methods such as leveraging large language models \cite{gao2024dag} to orchestrate unimanual policies can realize weak zero-shot generalization, but struggle with contact-rich tasks without learning. 

% \noindent\textbf{Historical Memory.}
% \method embeds a 'time' variable of the times the policy is called into the robot's proprioception. This provides a task completion process to the agent, which may help it distinguish different states even if the visual observations are the same, such as the before and after pressing the buttons in the \texttt{press buttons} tasks. However, this progressive memory contains rare information about the previous trials, which is of great importance for error recovery. Possible improvements include maintaining a memory bank that contains compressed information of historical frames~\cite{guhur2023instruction}.

% ==============================================
% In this supplementary material, we provide additional details and experiments not included in the main paper due to limitations in space.
% \begin{itemize}
%     \item \Cref{a:dataset}: Details of the RLBench dataset and the training pipeline used in our experiments.
%     \item \Cref{a:implementation_details}: Additional implementation details of our \method.
%     \item \Cref{a:additional_quantitative}: Supplementary quantitative analysis.
%     \item \Cref{a:additional_qualitative}: Supplementary qualitative analysis.
% \end{itemize}

% \section{Additional Implementation Details}

% \noindent \textbf{PerAct.} 

% \noindent \textbf{PerAct-LF.} 

% \noindent \textbf{PerAct$^2$.} 

\end{document}